\documentclass[10pt,twocolumn,letterpaper]{article}

\usepackage[pagenumbers]{cvpr} 

\definecolor{cvprblue}{rgb}{0.21,0.49,0.74}
\usepackage[pagebackref,breaklinks,colorlinks,allcolors=cvprblue]{hyperref}
\usepackage{xcolor}
\usepackage[font=small]{caption}
\usepackage{subcaption}
\usepackage{booktabs}
\usepackage{multirow}
\usepackage{makecell}
\usepackage{colortbl}
\usepackage{array}
\usepackage{caption}
\usepackage{adjustbox}
\usepackage{float}
\usepackage{placeins}
\usepackage{hyperref}
\usepackage{graphicx}   

\definecolor{orange}{RGB}{255,190,122}  
\definecolor{red}{RGB}{250,127,111}  
\definecolor{blue}{RGB}{130,176,210}  
\definecolor{purple}{RGB}{100,90,160} 
\definecolor{ArrowBlue}{RGB}{103,162,237}    
\definecolor{ArrowYellow}{RGB}{241,190,80}   
\definecolor{ArrowRed}{RGB}{220,103,90}      
\definecolor{ArrowGreen}{RGB}{121,183,110}   
\def\paperID{1014} 
\def\confName{CVPR}
\def\confYear{2026}

\title{%
\vspace{-1.2em}
  \raisebox{-0.2\height}{\includegraphics[height=4.2ex]{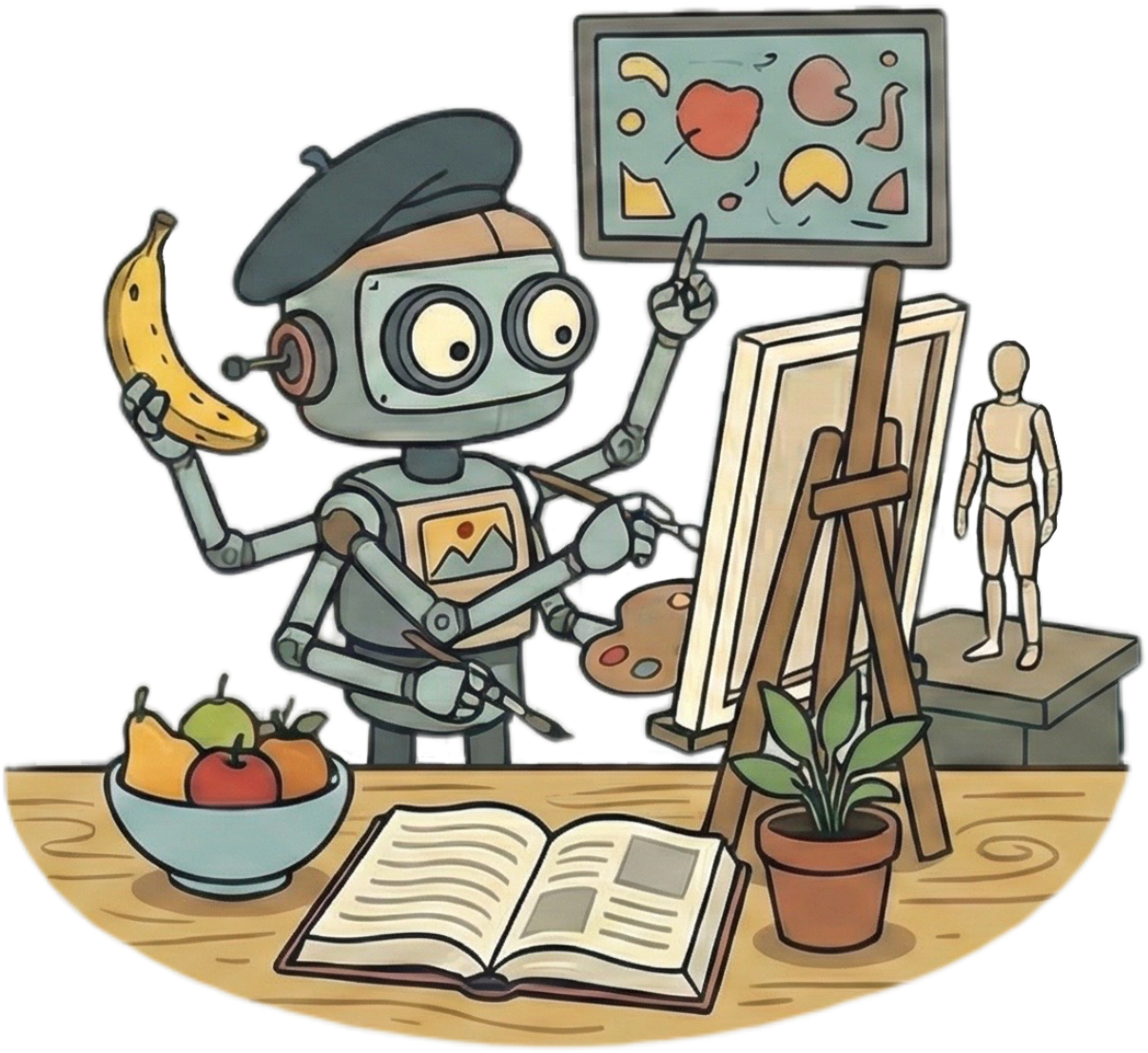}}\hspace{0.4em}%
  MICo-150K: A Comprehensive Dataset Advancing Multi-Image Composition
  \vspace{-1.2em}
}

\author{
\textit{Xinyu Wei}$^{1,5*}$, \textit{Kangrui Cen}$^{1,5*}$, \textit{Hongyang Wei}$^{2,5*}$, \textit{Zhen Guo}$^{1,5}$, \textit{Kai Cui}$^{3}$,\\
\textit{Bairui Li}$^{1,5}$, \textit{Zeqing Wang}$^{4,5}$, \textit{Jinrui Zhang}$^{1,5}$, \textit{Lei Zhang}$^{1,5\dagger}$\\[2pt]
\small
$^1$Hong Kong Polytechnic University\quad
$^2$Tsinghua University\quad
$^3$Peking University\quad
$^4$Sun Yat-Sen University\quad
$^5$OPPO Research Institute\quad
}

\begin{document}
\maketitle
\begingroup
\renewcommand\thefootnote{}\footnotetext{
$*$ Equal contribution. \quad
$\dagger$ Corresponding author.
}
\addtocounter{footnote}{0}
\endgroup
\begin{abstract}
In controllable image generation, synthesizing coherent and consistent images from multiple reference inputs, 
i.e., \textbf{Multi-Image Composition} (MICo), remains a challenging problem, partly hindered by the lack of high-quality training data.
To bridge this gap, we conduct a systematic study of MICo, categorizing it into 7 representative tasks and curate a large-scale collection of high-quality source images and construct diverse MICo prompts.
Leveraging powerful proprietary models, we synthesize a rich amount of balanced composite images, followed by human-in-the-loop filtering and refinement, resulting in \textbf{MICo-150K}, a comprehensive dataset for MICo with identity consistency.
We further build a Decomposition-and-Recomposition (De\&Re) subset, where 11K real-world complex images are decomposed into components and recomposed, enabling both real and synthetic compositions.
To enable comprehensive evaluation, we construct \textbf{MICo-Bench}, a curated benchmark of 897 cases across Object-Centric, Human-Centric, Human-Object Interaction, and De\&Re tasks, and further introduce a new metric, \textbf{Weighted-Ref-VIEScore}, specifically tailored for MICo evaluation.
Finally, we fine-tune multiple models on \textbf{MICo-150K} and evaluate them on \textbf{MICo-Bench}. 
The results show that MICo-150K effectively equips models without MICo capability and further enhances those with existing skills.
Notably, our baseline model, \textbf{Qwen-MICo}, fine-tuned from Qwen-Image-Edit, matches \textbf{Qwen-Image-2509} in 3-image composition while supporting arbitrary multi-image inputs beyond the latter’s limitation.
Our dataset, benchmark, and baseline collectively offer valuable resources for further research on Multi-Image Composition.
Project Page: \href{https://MICo-150K.github.io/}{\textcolor{blue}{https://MICo-150K.github.io/}}
\end{abstract}
    
\vspace{-1.0em}
\section{Introduction}
\label{sec:intro}
With the rapid advancement of generative AI, the capability of image synthesis has expanded dramatically. Both text-to-image (T2I)~\citep{rombach2022-latentdiff, openai2023dalle3, podell2023sdxl, chen2023pixal-alpha, li2024hunyuandit, cao2025hunyuanimage30, wu2025qwen-image, esser2024scalingrectifiedflowtransformers, blackforestlabs2024flux, gao2025seedream30technicalreport, gong2025seedream20nativechineseenglish, tong2025delvingrlimagegeneration, guo2025can, jiang2025t2i, guo2025thinking} and image-to-image (I2I)~\citep{brooks2023instructpix2pixlearningfollowimage, zhang2024magicbrushmanuallyannotateddataset, wang2025seededit30fasthighquality, deng2025emergingpropertiesunifiedmultimodal, labs2025flux1kontext, wu2025omnigen2explorationadvancedmultimodal, liu2025step1xeditpracticalframeworkgeneral, openai2025gpt4o_image, wang2025skyworkunipicunifiedautoregressive, wei2025skyworkunipic20building, lin2024pixwizard,pure} models can produce visually stunning and photorealistic results. 
Meanwhile, these models have possessed richer world knowledge and can support an increasingly diverse set of condition modalities~\citep{zhang2023addingconditionalcontroltexttoimage, mou2023t2iadapterlearningadaptersdig, qin2023unicontrolunifieddiffusionmodel}, ranging from complex textual prompts to depth maps, sketches, and reference images.
Among them, generating new images following the textual intent of users while maintaining identity (ID) consistency with reference images (particularly facial ID~\citep{ruiz2023dreamboothfinetuningtexttoimage, ruiz2024hyperdreamboothhypernetworksfastpersonalization, chen2024disenboothidentitypreservingdisentangledtuning, Toma_evi__2025, wang2024instantidzeroshotidentitypreservinggeneration, li2023photomakercustomizingrealistichuman, ye2023ipadaptertextcompatibleimage, cui2024idadapterlearningmixedfeatures, guo2024pulidpurelightningid, Cui_2024_CVPR}), a task often referred to as personalized generation~\citep{patel2024conceptbedevaluatingconceptlearning, patel2024lambdaeclipsemulticonceptpersonalizedtexttoimage, wang2025msdiffusionmultisubjectzeroshotimage} or in-context generation~\citep{labs2025flux1kontext, song2025insertanythingimageinsertion, wu2025lesstomoregeneralizationunlockingcontrollability, zhang2025incontexteditenablinginstructional}, represents one of the most valuable capabilities of modern generative models. 
Recent works such as FLUX.Kontext~\citep{labs2025flux1kontext} and Qwen-Image~\citep{wu2025qwen-image} have made remarkable progress in this direction. 
However, these systems only support a single-reference input, limiting the ability to integrate multiple entities into a unified and coherent composition, restricting users’ freedom of creative design.

\textbf{Multi-Image Composition} (MICo) with identity consistency, which involves combining content from multiple \textit{source images} into a coherent \textit{target image} while faithfully adhering to the textual prompts, stands as one of the most challenging yet underexplored tasks in generative image modeling.
Proprietary models like GPT-Image-1~\citep{sima2024gpt4oimage}, Nano-Banana~\citep{google2024nanobanana}, and Seedream 4.0~\citep{seedream2025seedream40nextgenerationmultimodal} have pushed this task to a new level of realism, coherence, and controllability.
However, open-source community remains a significant gap in MICo compared to proprietary models, partly due to the lack of high-quality datasets tailored for this task.

\begin{figure}[t!]
  \centering
  \includegraphics[width=\linewidth]{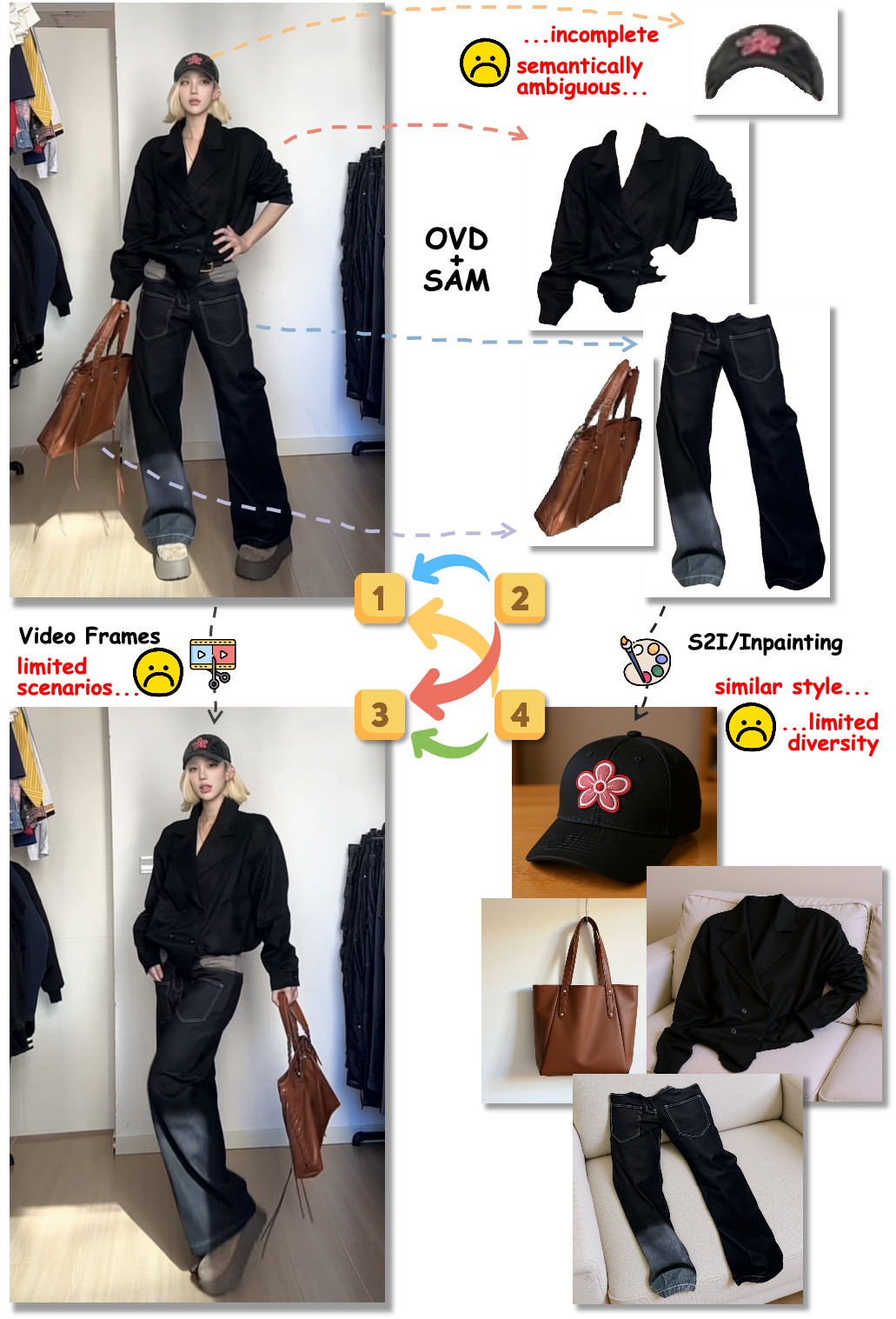}
    \caption{
Previous MICo methods typically collect high-quality images or video frames as \textbf{target images (1)}. Using Open-Vocabulary Detectors (OVD)~\citep{liu2024groundingdinomarryingdino} and SAM~\citep{kirillov2023SAM}, objects within targets are segmented to obtain \textbf{source images (2)}. Some methods enhance the targets by retrieving additional frames of the same subject from videos \textbf{(3)}, or enhance the sources using S2I (Subject-to-Image) or inpainting models \textbf{(4)}.
Training pairs are then constructed along multiple paths: \textbf{\color{ArrowBlue}(2→1)}, \textbf{\color{ArrowRed}(2→3)}, \textbf{\color{ArrowYellow}(4→1)}, and \textbf{\color{ArrowGreen}(4→3)}.
However, the masks in \textbf{(2)} are often incomplete and semantically ambiguous; the generated images in \textbf{(4)} tend to share similar styles, content, and limited diversity due to reliance on a few fixed generative models; the frames in \textbf{(3)} originate from a small number of high-quality videos, leading to limited scene variety and a lack of imaginative or complex multi-subject scenarios.
}
  \label{fig:other-data-wrong}
  \vspace{-1.0em}
\end{figure}

ConceptBed~\citep{patel2024conceptbedevaluatingconceptlearning} provides a large-scale dataset of ID-preserving images across varying concepts but does not include MICo data. Echo-4o~\citep{ye2025echo4o} introduces such data only as a small subset, with limited diversity in sources and prompts.
Some approaches achieve MICo without explicitly constructing composed targets \cite{xu2024versatilediffusiontextimages,patel2024lambdaeclipsemulticonceptpersonalizedtexttoimage,chen2025lamiclayoutawaremultiimagecomposition}.
Another line of research~\citep{ma2024subjectdiffusionopendomainpersonalizedtexttoimage}, as shown in Fig.~\ref{fig:other-data-wrong}, constructs training pairs by segmenting instances from full images using GroundingDINO~\citep{liu2024groundingdinomarryingdino} and SAM~\citep{kirillov2023SAM} 
as sources while originals as targets ~\citep{huang2024resolvingmulticonditionconfusionfinetuningfree,chen2025xverseconsistentmultisubjectcontrol}.
However, such paradigms often yield incomplete and semantically ambiguous samples.
Later methods such as UNO~\citep{wu2025lesstomoregeneralizationunlockingcontrollability}, OmniGen2~\citep{wu2025omnigen2explorationadvancedmultimodal}, and DreamO~\citep{mou2025dreamounifiedframeworkimage} enhance semantic richness by synthesizing new contexts; however, the weak generative backbones they used lead to highly homogeneous results in both style and content. 
Recently, DreamOmni2~\citep{xia2025dreamomni2multimodalinstructionbasedediting} introduces a three-stage data generation pipeline, achieving more controllable and consistent compositions. Nevertheless, the synthesized data still suffer from stylistic and contextual homogeneity.

In summary, existing MICo datasets face two main limitations: \textbf{(1)} many source or target images are generated by a few fixed T2I models, leading to homogeneous content and a noticeable quality gap compared to proprietary counterparts; \textbf{(2)} datasets based on real photos or video clips are limited in diversity, lacking imaginative scenarios, and biasing toward person-centric compositions, with insufficient coverage of object-centric or multi-subject cases.

\begin{figure*}[t!]
  \centering

  \begin{subfigure}{\textwidth}
    \centering
    \includegraphics[width=\textwidth]{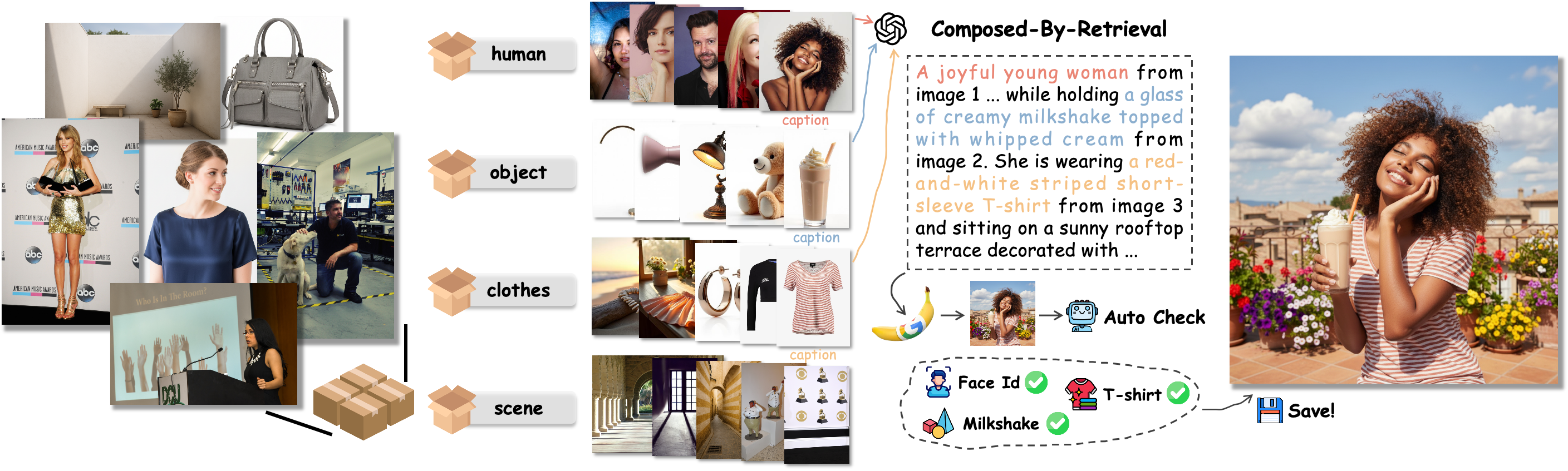}
    \caption{
    High-quality open-source data are collected and cleaned through a dedicated pipeline, categorized into four groups: \textbf{human}, \textbf{object}, \textbf{clothes}, and \textbf{scene}, each with detailed captions.
    For each task, a diverse set of source images is randomly sampled from these categories, and a multi-image composition prompt is generated by GPT-4o using image captions under a “\textbf{Composed-by-Retrieval}” strategy.
    The generated prompt is then fed into Nano-Banana \cite{google2024nanobanana} to synthesize composite images.
    Each output image undergoes an automated verification process including QwenVL2.5-72B~\citep{bai2025qwen25vltechnicalreport} and ArcFace~\citep{Deng_2022} to ensure that all source images are correctly represented in the final composition before being included in the dataset.
    }
    \label{fig:stacked:a}
  \end{subfigure}

  \vspace{4pt}

  \begin{subfigure}{\textwidth}
    \centering
    \includegraphics[width=\textwidth]{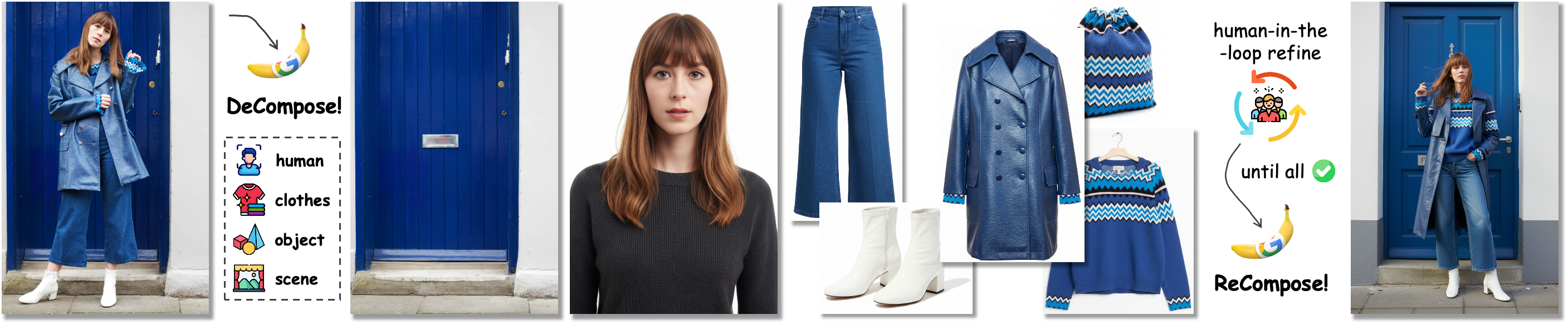}
    \caption{
We collect a large number of high-quality single-person portraits through our data-cleaning pipeline and use \textbf{Nano-Banana}~\citep{google2024nanobanana} to \textbf{decompose} each image into its constituent components—\textbf{scene}, \textbf{human}, \textbf{objects}, and \textbf{clothes}.
Human annotators then carefully inspect all decomposed components to ensure their quality.
Once all parts meet the required standards, \textbf{Nano-Banana} is used again to \textbf{recompose} them into a complete image.
}
    \label{fig:stacked:b}
  \end{subfigure}

  \caption{Construction pipeline of MICo-150K.
\textbf{(a)} The data construction pipeline for the \textbf{Human-Centric}, \textbf{Object-Centric}, and \textbf{HOI (Human–Object Interaction)} tasks.
\textbf{(b)} The pipeline for the \textbf{De\&Re (Decompose and Recompose)} task.
}
  \label{fig:data-pipeline}
\end{figure*}

To address these limitations, we introduce \textbf{MICo-150K}, a high-quality and comprehensive dataset specifically designed for multi-image composition tasks.
We first define a taxonomy with \textbf{3 major categories}, \textbf{7 sub-tasks}, and \textbf{27 fine-grained types}, along with a realistic \textbf{Decompose and Recompose (De\&Re) track}, as illustrated in Fig.~\ref{fig:data-pipeline}.
For each task, we design dedicated data collection and cleaning pipelines to curate high-quality source images of objects, humans, clothing, and scenes. Each image is paired with a detailed, descriptive caption aligned with its visual content.
To generate composition prompts, we use GPT-4o~\citep{openai2025gpt4o_image} with a \textbf{Compose-by-Retrieval} strategy, which combines compatible source captions to form rich and coherent prompts for each composition type.
These prompts are then fed into Nano-Banana~\citep{google2024nanobanana}, a state-of-the-art proprietary model, to synthesize high-quality composite images.
Finally, a human-in-the-loop post-filtering process is applied to retain only the most accurate and visually consistent results, yielding the final MICo-150K dataset.


It is also observed that there lacks a dedicated benchmark for evaluating MICo performance.
To fill this gap, we introduce \textbf{MICo-Bench}, a comprehensive benchmark comprising 897 carefully curated cases across four task groups: 138 Object-Centric, 168 Human-Centric, 291 Human-Object Interaction, and 300 challenging De\&Re cases. 
Each case is rigorously validated by human annotators to ensure diverse semantics and scene coverage across source images and text prompts.
For evaluation, we propose a new metric \textbf{Weighted-Ref-VIEScore}, a reliable and human-aligned metric for assessing MICo performance.

We fine-tune several T2I models, {BAGEL}~\citep{deng2025bagel}, {BLIP3-o}~\citep{chen2025blip3ofamilyfullyopen}, {Lumina-DiMOO}~\citep{xin2025luminadimooomnidiffusionlarge}, {OmniGen2}~\citep{wu2025omnigen2explorationadvancedmultimodal}, and {Qwen-Image-Edit}~\citep{wu2025qwen-image} on \textbf{MICo-150K} and evaluate them using \textbf{MICo-Bench}.
BLIP3-o and Lumina-DiMOO that originally lacked multi-image composition capabilities acquire strong MICo abilities after fine-tuning.
BAGEL and Qwen-Image-Edit, though not explicitly trained for multi-image composition, demonstrate emergent MICo capability that is significantly enhanced after fine-tuning on MICo-150K.
OmniGen2, which already possesses robust multi-image consistency, also exhibits further improvement after training.
Among these, the model fine-tuned from Qwen-Image-Edit, termed \textbf{Qwen-MICo}, serves as our primary baseline.
In summary, our contributions are threefold:
\textbf{(i)} We introduce \textbf{MICo-150K}, a large-scale high-quality dataset specifically designed for multi-image composition with identity consistency, and provide comparisons between real-world and synthetic compositions.;
\textbf{(ii)} We present \textbf{MICo-Bench} along with the \textbf{Weighted-Ref-VIEScore}, establishing a comprehensive and reliable evaluation framework for MICo task;
\textbf{(iii)} We observe that some strongly pretrained I2I models \textbf{naturally exhibit emergent MICo capabilities}. Extensive experiments further demonstrate the robustness and generality of MICo-150K, as models with diverse architectures and initialization states consistently achieve substantial improvements after fine-tuning.
\textbf{(iv)} We propose \textbf{Qwen-MICo} as a baseline. Trained solely on MICo-150K, it outperforms Qwen-Image-2509~\citep{qwen2025qwenimageedit2509} on MICo-Bench while lifting the restriction to support ($>3$) references.

\vspace{-0.2em}
\section{Related Work}
\label{sec:related_work}

\noindent\textbf{Training-free MICo Methods}.
Versatile Diffusion~\citep{xu2024versatilediffusiontextimages}, trained solely on standard I2I data, adopts a unified multi-flow diffusion framework optimized across diverse tasks, exhibiting certain MICo capabilities.
$\lambda$-ECLIPSE~\citep{patel2024lambdaeclipsemulticonceptpersonalizedtexttoimage} learns the CLIP latent space using image–text interleaved data and achieves MICo through a frozen diffusion inference backend without any specific supervision.
MIP-Adapter~\citep{huang2024resolvingmulticonditionconfusionfinetuningfree} introduces lightweight, learnable adapters into a frozen diffusion model, enabling MICo by training a small adapter.
LAMIC~\citep{chen2025lamiclayoutawaremultiimagecomposition} extends single-reference T2I models in a fully training-free manner by introducing \textit{Group Isolation Attention} and \textit{Region-Modulated Attention}, which support layout-aware MICo without modifying the underlying MMDiT parameters or requiring additional training data.


\vspace{+1mm}
\noindent\textbf{Trainable MICo Methods}.
Subject Diffusion~\citep{ma2024subjectdiffusionopendomainpersonalizedtexttoimage} pioneers a data construction paradigm that extracts background-free instances using GroundingDINO~\citep{liu2024groundingdinomarryingdino} and SAM~\citep{kirillov2023SAM} as source images, with the original images serving as targets. This strategy has provided valuable insights for building MICo datasets.
MS-Diffusion~\citep{wang2025msdiffusionmultisubjectzeroshotimage} follows this approach but constructs semantically aligned image pairs from different frames of the same object in video clips, avoiding direct copy–paste duplication.
UNO~\citep{wu2025lesstomoregeneralizationunlockingcontrollability} enhances model capability through progressive cross-modal alignment. It follows Subject Diffusion~\citep{ma2024subjectdiffusionopendomainpersonalizedtexttoimage} but employs a Subject-to-Image (S2I) model to generate new source images rather than using the original instances directly.
DreamO~\citep{mou2025dreamounifiedframeworkimage} introduces\textit{ feature routing constraints} into the DiT training process to ensure identity consistency. It uses PuLID~\citep{guo2024pulidpurelightningid} to generate multiple portraits of the same person ID and selects different frames of the same subject from videos for body-level data.
OmniGen2~\citep{wu2025omnigen2explorationadvancedmultimodal} employs two distinct decoding pathways for text and image with unshared parameters and achieves strong MICo ability through large-scale training. Its data pipeline also leverages multi-frame videos but includes refined filtering and inpainting strategies to restore background context.
XVerse~\citep{chen2025xverseconsistentmultisubjectcontrol} introduces a \textit{T-Mod Resampler} that processes each source image and injects its representation into a per-token modulation adapter, while incorporating VAE features to preserve fine-grained details.
DreamOmni2~\citep{xia2025dreamomni2multimodalinstructionbasedediting} proposes a novel three-stage data construction pipeline. It first introduces \textit{dual-branch attention mixing} in a T2I model to generate two semantically aligned images with partially shared features, then trains an extraction model on them, which is subsequently used to generate reference images.

\begin{table*}[!htbp]
  \centering
  \caption{
    Statistics of the \textbf{MICo-150k} dataset. 
    The dataset is carefully balanced to ensure diversity across task types and input images. 
    In the \textbf{Types} column, abbreviations denote component categories: 
    O = Object, S = Scene, P = Person, M = Man, W = Woman, and C = Cloth.
    }
    \vspace{-2mm}
  \label{tab:dataset_stats}
  \newcolumntype{L}[1]{>{\raggedright\arraybackslash}p{#1}}
\newcolumntype{C}[1]{>{\centering\arraybackslash}p{#1}}
\newcommand{\catcell}[2]{\cellcolor{#1!22}\textbf{#2}}
\newcommand{\tripleshade}[4]{ & \cellcolor{#1} #2 & \cellcolor{#1} #3 & \cellcolor{#1} #4 \\}

\renewcommand{\arraystretch}{0.9}
\setlength{\extrarowheight}{-1pt}
\setlength{\tabcolsep}{5pt}
\renewcommand{\arraystretch}{0.7}

\adjustbox{max width=\linewidth}{%
\begin{tabular}{L{4.4cm} L{4.1cm} L{4.4cm} L{4.4cm}}
\toprule
\textbf{Task} & \textbf{Sub-task} & \textbf{Types} & \textbf{Count} \\
\midrule

\multirow{2}{*}{\textbf{Object Centric}}
  \tripleshade{orange!22}{Object + Scene}{1O1S, 2O1S}{5014, 4999}
\cmidrule[\lightrulewidth](lr){2-4}
  \tripleshade{orange!22}{Object + Object}{2O, 3O, 4O, 5O}{10007, 10012, 5001, 4998}
\midrule

\multirow{2}{*}{\textbf{Person Centric}}
  \tripleshade{red!22}{Person + Person}
    {\makecell[l]{1M1W, 2M1W, 2W1M\\ 2M, 2W, 3M, 3W}}
    {\makecell[l]{3001, 3003, 3066\\ 2996, 3006, 2999, 2991}}
\cmidrule[\lightrulewidth](lr){2-4}
  \tripleshade{red!22}{Person + Scene}{1P1S, 2P1S}{4986, 4994}
\midrule

\multirow{3}{*}[-1.6ex]{\textbf{Human Object Interaction}}
  \tripleshade{blue!22}{Person + Objects}{1P1O, 1P2O, 2P1O, 2P2O}{5020, 4973, 5071, 5050}
\cmidrule[\lightrulewidth](lr){2-4}
  \tripleshade{blue!22}{Person + Clothes}{1P1C, 1P2C, 1P3C, 1P4C}{7031, 7017, 6942, 6973}
\cmidrule[\lightrulewidth](lr){2-4}
  \tripleshade{blue!22}{\makecell[l]{Person + Objects + Clothes}}
    {\makecell[l]{1P1C1O, 1P1C2O\\ 1P2C1O, 1P2C2O}}
    {\makecell[l]{5034, 5022\\ 5025, 5011}}
\midrule

\textbf{De\&Re}
  & \multicolumn{2}{>{\columncolor{purple!22}}c}{\textbf{Adaptive}}
  & \cellcolor{purple!22}{11677} \\
\bottomrule
\end{tabular}%
}

\end{table*}

\vspace{-0.2em}
\section{MICo-150K Dataset}
\label{sec:mico150k}
\begin{figure*}[t!]
  \centering
  \includegraphics[width=0.999\textwidth]{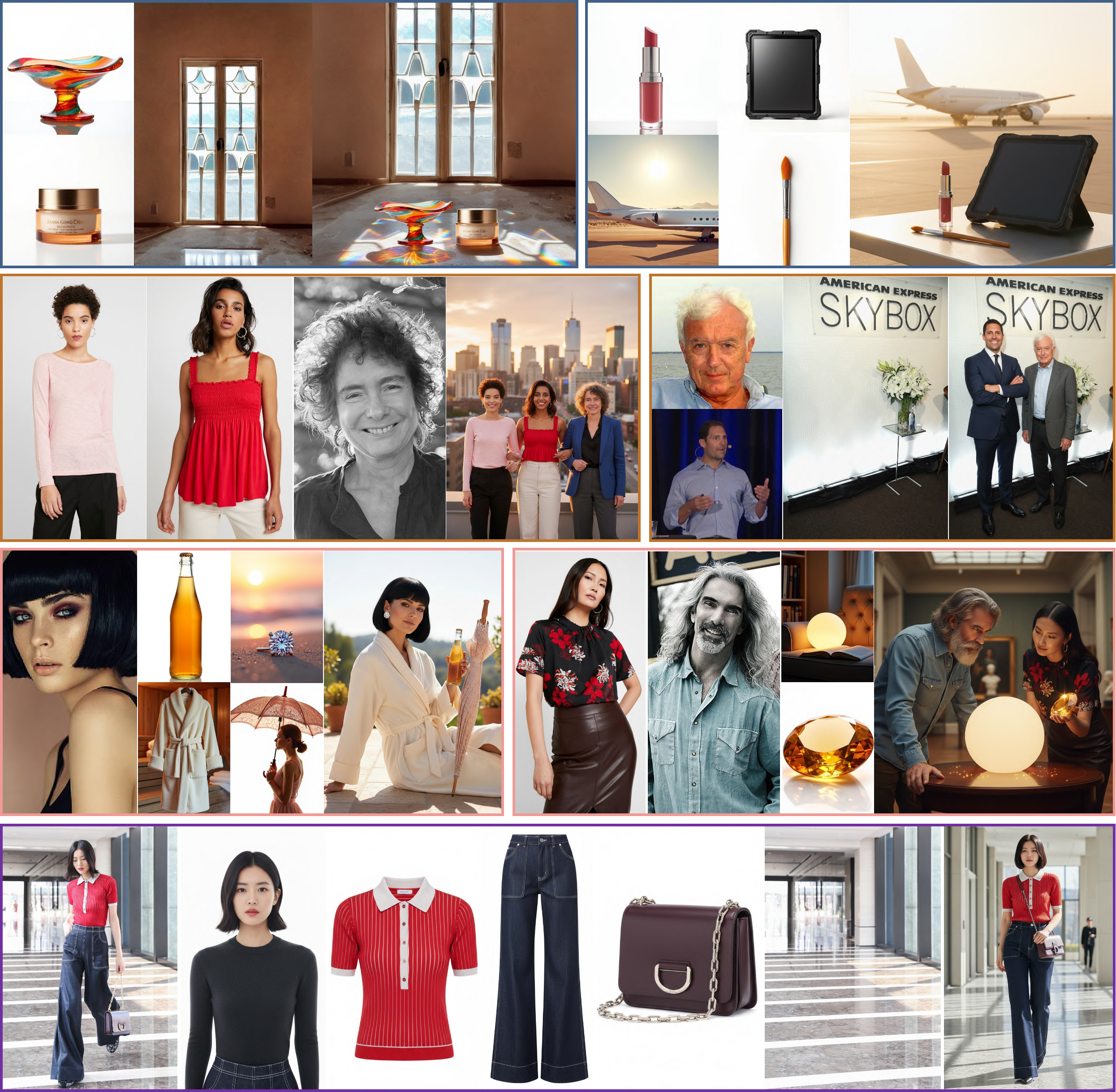}
  \caption{
Visualization examples from the MICo-150K dataset. 
\textbf{Row 1 \textcolor{blue}{(Object-Centric)}}: “2 objects + scene” and “4 objects” compositions. 
\textbf{Row 2 \textcolor{orange}{(Person-Centric)}}: “3 women” and “2 persons + scene”. 
\textbf{Row 3 \textcolor{red}{(Human-Object Interaction)}}: “1 person + 4 objects” and “2 persons + 2 objects”. 
\textbf{Row 4 \textcolor{purple}{(De\&Re)}}: the first image is a real-world photo, the last is the recomposed result, with intermediate visual elements including decomposed persons, objects, clothes, and scene components.
}
  \label{fig:dataset-case}
\end{figure*}

This section provides the detailed construction pipeline of the \textbf{MICo-150K}. 
Sec.~\ref{sec:source_image} describes the collection of source images, while 
Sec.~\ref{sec:MICO_tasks} details the construction processes of the four task categories. For clarity, the number of cases for each sub-task is summarized in Tab.~\ref{tab:dataset_stats}, representative examples from each task are illustrated in Fig.~\ref{fig:dataset-case}.
All image licensing and usage permissions are detailed in Sec.~\ref{app:Reproducibility}.

\subsection{Source Image Collection}
\label{sec:source_image}

\noindent\textbf{Object Images}.
We collected object images and the corresponding captions from the Subject200k~\citep{tan2025ominicontrolminimaluniversalcontrol} dataset. 
To ensure quality, we employed \textbf{Qwen2.5-VL-72B}~\citep{bai2025qwen25vltechnicalreport} for filtering, removing images containing humans, obvious blur, corruption artifacts, or ambiguous semantics. 
Each retained image was assigned a detailed, descriptive caption and category label. 
Within each category, we extracted visual features using \textbf{DINO-v3}~\citep{siméoni2025dinov3} and \textbf{SigLIP2}~\citep{tschannen2025siglip2multilingualvisionlanguage}, concatenated them, and applied \textbf{DBSCAN}~\citep{Ester1996ADA} clustering to eliminate redundancy, keeping only one representative image per visual-semantical cluster. 
After this rigorous cleaning process, we obtained \textbf{31.5K} high-quality object images.

\vspace{+2mm}
\noindent\textbf{Human Images}.
We collected 13.7K high-quality model images from the VITON-HD~\citep{choi2021vitonhdhighresolutionvirtualtryon} try-on dataset, 6K upper-body portraits from the Headshot iStockPhoto dataset~\citep{bkm1804_headshot_istockphoto}, 3K high-resolution portraits from the Headshot Pexels v1 dataset~\citep{bkm1804_headshot_pexels_v1}, professional portrait photos from the X2I-Subject-Driven dataset~\citep{xiao2024omnigenunifiedimagegeneration}, and 144K photos of 5,403 celebrities. 
We then employed \textbf{Qwen2.5-VL-72B}~\citep{bai2025qwen25vltechnicalreport} to remove images containing multiple faces, back views, or heavily occluded faces. 
Each remaining image was annotated with a detailed caption describing gender, attributes, clothing, and posture. 
After the cleaning and annotation process, we obtained \textbf{44.6K} high-quality human portraits, ensuring maximal diversity in facial identities.
A detailed analysis of human-face images is provided in Sec.~\ref{app:Human-Face}.

\vspace{+2mm}
\noindent\textbf{Cloth Image}.
We collected clothing images from the Subject200k~\citep{tan2025ominicontrolminimaluniversalcontrol} and the VITON-HD~\citep{choi2021vitonhdhighresolutionvirtualtryon} try-on dataset.
The clothing items were categorized by type (upper garments, lower garments, shoes, accessories) and gender, forming $4_{type}*2_{gender}=8$ subcategories in total.
We used Qwen2.5-VL-72B~\citep{bai2025qwen25vltechnicalreport} to filter out unclear or inappropriate items and generate descriptive captions for each image.
The final source pool contains \textbf{17.3K} upper garments, \textbf{1.3K} pants, \textbf{428} pairs of shoes, and \textbf{7.8K} accessories.

\vspace{+2mm}
\noindent\textbf{Scene Images}.
We collected scene images from the Mulan~\citep{tudosiu2024mulanmultilayerannotated}, Echo-4o~\citep{ye2025echo4o}, and SUN397~\citep{5539970} datasets. However, the overall quality of these scenes varies significantly.
To make them suitable, we first filtered out images whose shorter side was below 512 pixels. For the remaining images, we used Qwen2.5-VL-72B~\citep{bai2025qwen25vltechnicalreport} to perform a series of quality checks: 
\textit{(1)} the scene must be open and contain no people; 
\textit{(2)} it should depict a realistic, natural environment with no visible artifacts; 
\textit{(3)} it should be clean and well-organized, without elements such as toilets or garbage; 
\textit{(4)} it must not be overly confined and should be spacious enough to accommodate at least one person.
In addition, we used GPT-4o~\citep{openai2024gpt4ocard} to generate extra portrait-friendly scenes. We manually collected 200 high-frequency scene descriptions from AIGC community forums used for portrait generation, assigned concise text labels to them, and then prompted GPT-4o to synthesize the corresponding scenes.
After all filtering and generation, we obtained a final set of \textbf{11K} high-quality scenes suitable for the MICo tasks.

\vspace{+2mm}
\noindent\textbf{Compose-by-Retrieval}.
We employ Nano-Banana~\citep{google2024nanobanana} to compose the source images according to task settings, producing high-quality composite results.
However, directly sampling source images at random from the pools may lead to semantically incompatible combinations, for instance, pairing a male athlete with high-heeled shoes, resulting in poor generation quality.
To address this, we propose the \textbf{Compose-by-Retrieval} strategy.
After choosing a primary (subject) image, we sample multiple candidates for clothing, scenes, or objects from respective pools.
We then provide the subject image, candidate images, and their detailed captions as the context to GPT-4o~\citep{openai2024gpt4ocard} to select the most semantically compatible combination for composition.

\begin{figure}[t]
  \centering
  \begin{subfigure}{\linewidth}
    \centering
    \includegraphics[width=0.9\linewidth]{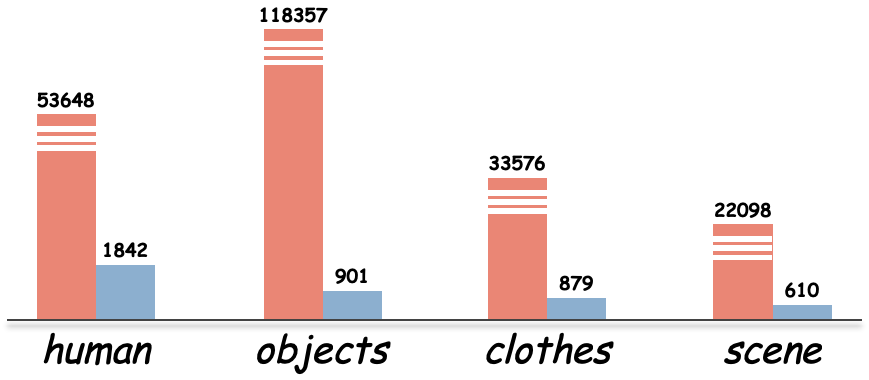}
    \caption{Statistics of source image counts across different categories.}
  \end{subfigure}

  \vspace{4pt} 

  \begin{subfigure}{\linewidth}
    \centering
    \includegraphics[width=0.9\linewidth]{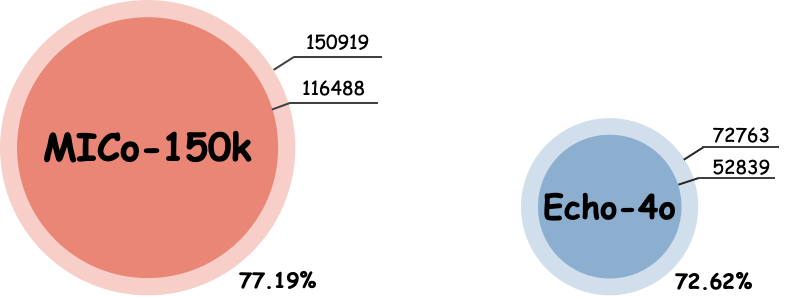}
    \caption{Semantic redundancy in text prompts. We extract text embeddings using CLIP and set a cosine similarity threshold of 0.85 to identify duplicates.}
  \end{subfigure}
\vspace{-5mm}
  \caption{High-quality multi-image composition datasets that are non-segmentation-based and not generated by Flux series~\citep{labs2025flux1kontext} are extremely rare; to the best of our knowledge, only Echo-4o~\citep{ye2025echo4o} is publicly available. MICo-150K significantly surpasses it in both source image diversity and text prompt semantic variety.}
  \label{fig:barpie}
  \vspace{-5mm}
\end{figure}

\subsection{Construction of MICo Tasks}
\label{sec:MICO_tasks}
\noindent\textbf{Object-Centric Tasks}.
We define two sub-tasks for object-centric composition.
The \textit{Object + Object} task involves combinations of one to four distinct objects, where source images are randomly sampled for composition.
The \textit{Object + Scene} task covers compositions of a single object or two objects placed within a specific scene. We adopt the proposed \textit{Compose-by-Retrieval} strategy, which provides multiple candidate scenes for each object and adaptively selects the most semantically compatible one.
Previous MICo methods typically use source image captions directly as prompts, such as ``\texttt{\textit{Combine 2 images according to \textless Caption~A\textgreater{} and \textless Caption~B\textgreater{}.}}"
To promote prompt diversity and naturalness, we instead treat source captions as contextual input for GPT-4o, which generates a more coherent composition prompt.
Furthermore, GPT-4o annotates the prompt with explicit token-to-source mappings, providing a foundation for future studies on latent space alignment.
We then used Nano-Banana to synthesize composite images and employed Qwen2.5-VL-72B for verification.
Each source image is paired with the generated target, and the model checks whether the objects from the source appear correctly in the final composition.

\vspace{+1mm}
\noindent\textbf{Person-Centric Tasks}.
We define two sub-tasks: \textit{Person + Person}, which covers 7 different gender and group combinations, and \textit{Person + Scene}, which composes 1 or 2 individuals placed within specific scenes.
We find that omitting explicit gender specification in prompts degrades generation quality. Therefore, we explicitly state gender information in the \textit{Person + Person} task.
For the \textit{Person + Scene} task, we adopt the Compose-by-Retrieval strategy to select semantically compatible scenes for each person.
During quality control, objects and scenes are verified using Qwen2.5-VL-72B, while face ID consistency is evaluated via ArcFace~\citep{Deng_2022} embeddings. Specifically, we extract face ID embeddings from both the source and generated images and compute the optimal matching using the Hungarian algorithm~\citep{Kuhn1955TheHM}. A generation is considered valid only if all matched pairs exceed a task-specific similarity threshold.

\vspace{+1mm}
\noindent\textbf{Human Object Interaction Tasks}.
We define three sub-tasks: \textit{Person + Object}, \textit{Person + Clothes}, and \textit{Person + Object + Clothes}, each further divided into four task variants as summarized in Tab.~\ref{tab:dataset_stats}(third row).
During the composition phase, we explicitly specify the gender of the human source and apply the Compose-by-Retrieval Strategy to select semantically compatible objects or clothes for each case.
For quality verification, we employ both Qwen2.5-VL-72B and ArcFace. The former ensures that all visual entities are correctly rendered, while the latter extracts face ID and evaluates the consistency with the source images.

\vspace{+1mm}
\noindent\textbf{Decompose-and-Recompose (De\&Re) Tasks}.
This task is the most complex and comprehensive part in MICo-150K.
We first curated a large collection of high-quality single-person portraits from the CC12M~\citep{changpinyo2021conceptual12mpushingwebscale} dataset.
Using Nano-Banana, we decomposed these complex, realistic photographs into separate components such as humans, clothing, objects, and scenes. Human annotators then conducted fine-grained quality checks, identifying failure cases, such as those missing object identities, or directly copy-pasted, lacking sufficient variance.
Then, the annotators rewrote component-specific prompts to correctly extract the intended elements.
After this human-in-the-loop refinement process, Nano-Banana was used again to recombine them into complete images.
Thus, each set of components produces two versions: a real-world composition and a recomposed synthesized one, as illustrated in Fig.~\ref{fig:data-pipeline}.

After the above construction process, the statistics of the dataset’s source images and text prompts are shown in Fig.~\ref{fig:barpie}, demonstrating significantly greater diversity compared to other datasets~\citep{ye2025echo4o} with similar functionality.

\begin{figure*}[h!]
  \centering
  \includegraphics[width=0.999\textwidth]{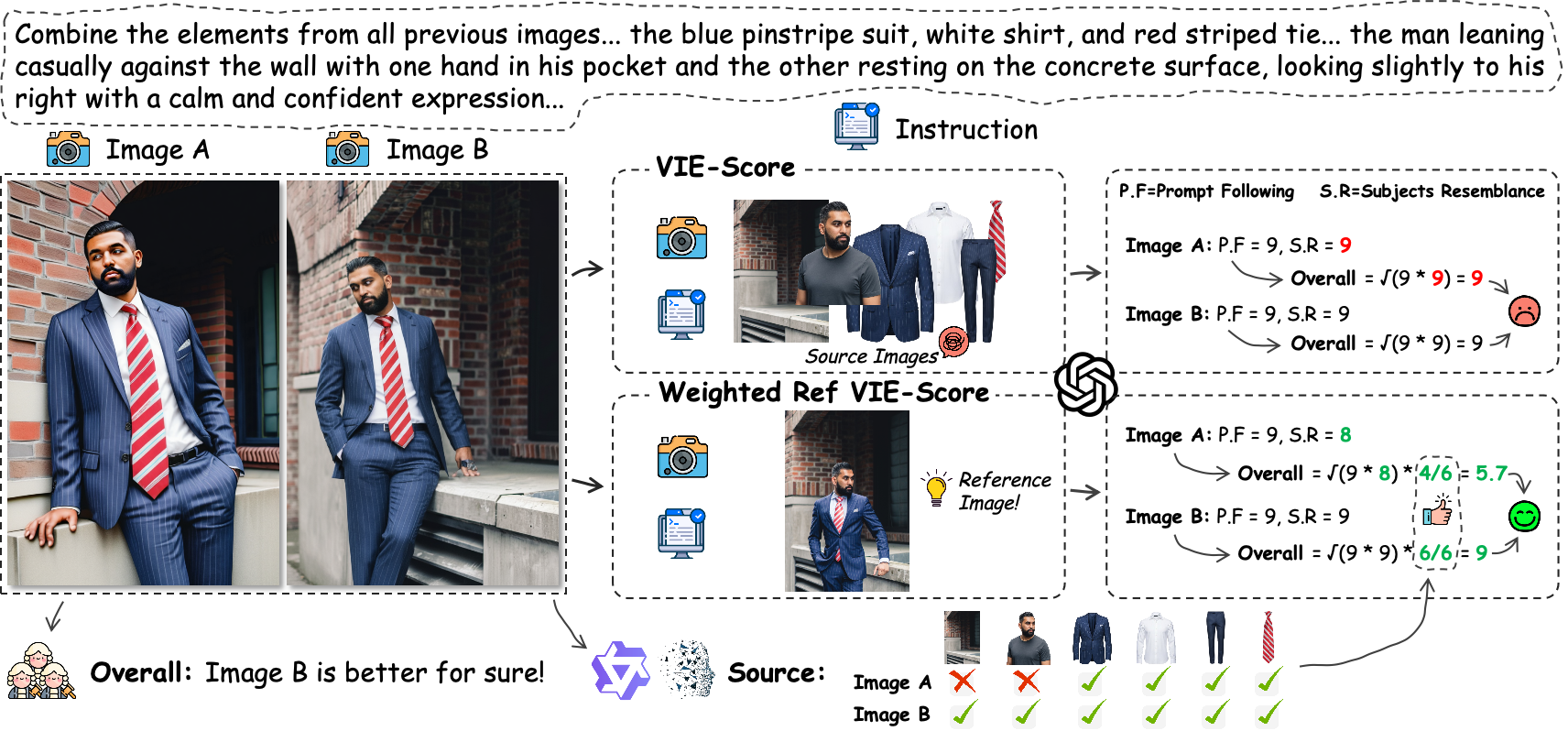}
  \caption{Traditional VIEScore requires inputting all source and generated images into the evaluator, which often leads to degraded performance as \textbf{the VLM evaluator's cross-image attention becomes overloaded}. This prevents the model from fully understanding each image and accurately determining whether every source appears in the target, resulting in substantial scoring errors (in this example, all the three human evaluators unanimously agreed that Image B was far superior). In contrast, MICo-Bench first assesses whether each source image appears in the generated result to produce weights. Each case also includes a verified reference image that contains all sources. During evaluation, the VLM judge compares only the generated image and the reference image, enabling human-level judgment accuracy.}
  \label{fig:bench}
\end{figure*}

\section{MICo-Bench}
\label{sec:mico-bench}
We first analyze why conventional image generation metrics are insufficient for MICo task (Sec.~\ref{sec:howto}), then introduce the Weighted-Ref-VIEScore (Sec.~\ref{sec:WeightedRefVIEScore}). Finally, we describe the curation pipeline of our benchmark (Sec.~\ref{sec:benchcuration}).

\subsection{Why Existing I2I Metrics Fail for MICo?}
\label{sec:howto}
With the rapid rise of general-purpose Vision-Language Models (VLMs), evaluating generative tasks such as I2I and T2I through VLMs has become increasingly popular.
VIEScore~\citep{ku2024viescoreexplainablemetricsconditional} offers a representative framework for VLM-based image evaluation by decomposing overall quality into two dimensions: Semantic Consistency (\textbf{SC}) and Perceptual Quality (\textbf{PQ}).
The final score is computed as $SC \times PQ$.
OmniContext Bench~\citep{wu2025omnigen2explorationadvancedmultimodal} follows this framework and employs GPT-4o~\citep{openai2024gpt4ocard} to evaluate the \textbf{SC} score of composition results in terms of Prompt Following (\textbf{PF}) and Subject Resemblance (\textbf{SR}).
However, it has a key limitation: it requires all source images to be input simultaneously during evaluation. Although modern VLMs like GPT-4o~\citep{openai2024gpt4ocard} are highly capable, their \textbf{cross-image attention remains limited}. When too many images are provided, the model struggles to accurately perceive each image’s content, leading to unreliable assessments of composition quality and consequently wrong scores, as illustrated in first row of Fig.~\ref{fig:bench}.

\subsection{Weighted-Ref-VIEScore}
\label{sec:WeightedRefVIEScore}
To overcome the limitations of existing evaluation frameworks, we introduce a weighted, reference-based evaluation paradigm tailored for multi-image composition.

\vspace{+1mm}
\noindent\textbf{Weighting}:
During each evaluation, every non-human source image is first paired with the generated result and fed into a VLM to determine whether the source element appears in the generated output.
In our released evaluator, we use Qwen3-VL-30B-A3B-Instruct for this binary verification.
For facial source images, we instead employ ArcFace~\citep{Deng_2022}; the cosine similarity is mapped to a graded preservation score rather than a hard binary decision.
The final visibility coefficient $W$ is the average preservation score over all source elements. However, relying solely on these weights may be exploited by models, for instance, \textbf{by trivially copy-pasting all sources into the output}. To prevent such bias, we incorporate a complementary reference-based mechanism.

\vspace{+1mm}
\noindent\textbf{Reference}:
For each case, we first generate a high-quality reference image using Nano-Banana~\cite{google2024nanobanana}, followed by rigorous human verification to ensure that all source elements are faithfully represented, with the necessary visual variance and semantic alignment with the given prompt.
During evaluation, GPT-5.4 compares each generated image with its reference image individually, rather than jointly processing all source images, as illustrated in Fig.~\ref{fig:bench} (Weighted-Ref-VIEScore).
This design yields a more accurate and human-aligned \textbf{SC} score.
The \textbf{PQ} score is then computed solely based on the generated image.
Finally, the overall score is defined as $\text{Score}=W \times \text{SC} \times \text{PQ}$, where $\text{SC}=\sqrt{SR \times PF}$ and $\text{PQ}$ is the geometric mean of the perceptual-quality sub-scores.
Extensive user studies demonstrate that our metric aligns with human preference significantly better than existing alternatives.

A more detailed computation procedure is provided in Sec.~\ref{app:Weighted-Ref-VIEScore}.
The human study validating the Weighted-Ref-VIEScore is presented in Sec.~\ref{app:humanstudy}.
Additional ablation studies, including \textit{Reference Image and Style Homology Risk}, \textit{VLM Evaluator–Generator Coupling}, and \textit{Resistance to Copy-Paste Attacks}, are discussed in Sec.~\ref{app:AdditionalExperiment}.

\subsection{Benchmark Curation}
\label{sec:benchcuration}
We construct MICo-Bench by selecting 897 representative cases from MICo-150K: 138 Object-Centric cases, 168 Human-Centric cases, 291 Human-Object Interaction cases, and 300 De\&Re cases.
Each case includes source images, a text prompt, a human-verified reference image, and indices that identify human-face sources for ArcFace-based weighting.
To ensure annotation reliability and fairness, we invited three independent reviewers who were not involved in this project to perform the selection. These reviewers were unaware of the distinctions between the general composition tasks and the De\&Re task. Each case was evaluated from two perspectives:
(1) whether all objects from the source images were correctly preserved in the generated result, and
(2) whether the final image aligned with the textual instruction.
Only the cases unanimously judged as satisfactory by all three reviewers were retained for the final benchmark.

\begin{figure}[t]
  \centering
  \includegraphics[width=\linewidth]{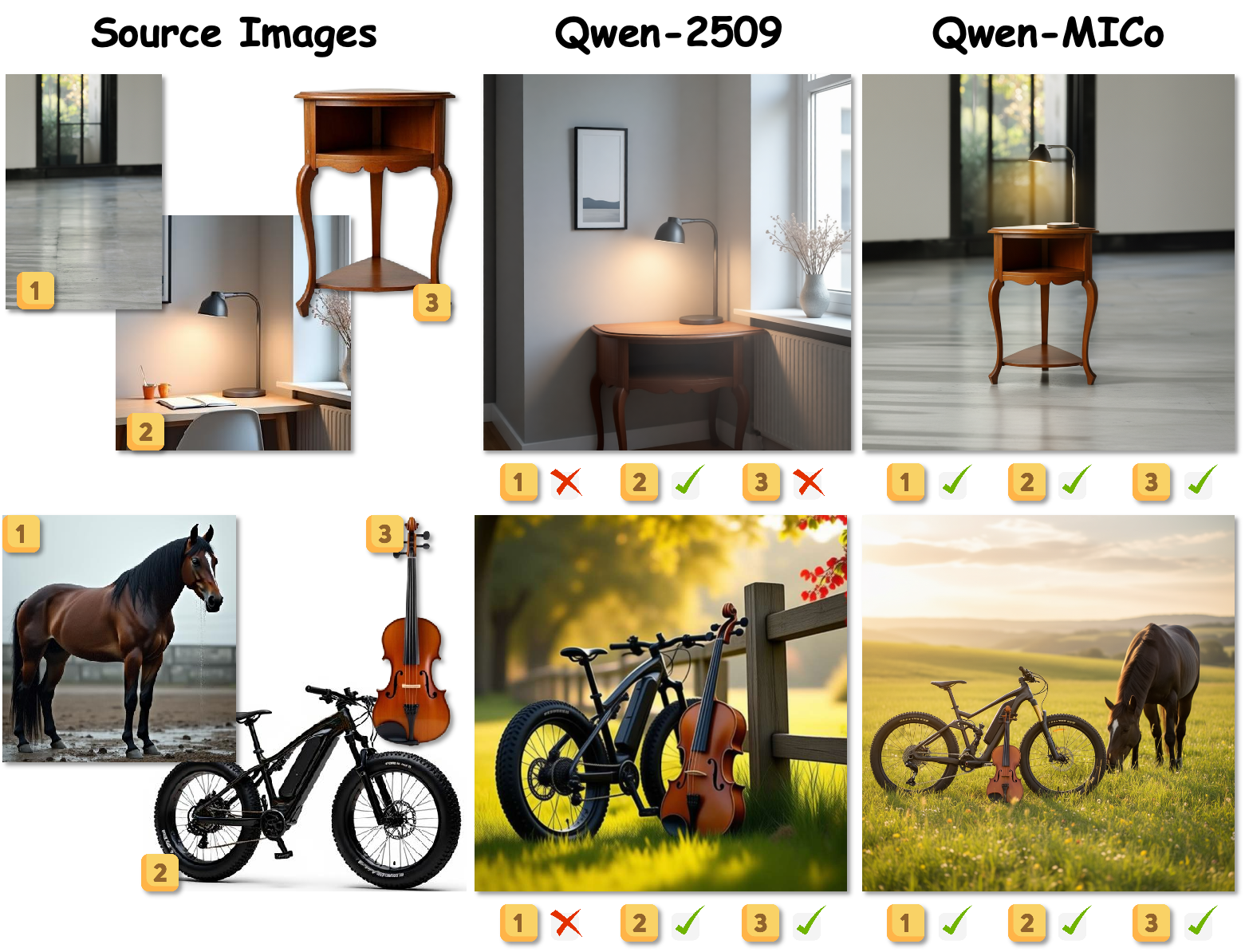}
  \vspace{-5mm}
  \caption{Qwen-Image-2509 is trained on a massive-scale dataset and supports only three-image inputs, while Qwen-MICo is trained solely on MICo-150K and supports arbitrary multi-image inputs, producing images with higher aesthetic quality.}
  \label{fig:qwen-mico}
\end{figure}
\begin{table*}[t!]
  \centering
  \caption{
    Performance comparison on the released MICo-Bench across closed-source systems, original open-source models, and MICo-finetuned models.
    Scores are computed with Weighted-Ref-VIEScore on 897 benchmark cases; the best result in the closed-source group and the best result in the open-source/MICo-finetuned group are highlighted in \textbf{bold}.
  }
  \label{tab:mico_bench_results}
  \vspace{-2mm}
  \newcommand{\thinrule}{\textcolor{black!60}{\vrule width 0.03pt}}
\newcolumntype{!}{@{\hspace{4pt}\thinrule\hspace{4pt}}}

\small
\renewcommand{\arraystretch}{1.25}
\setlength{\tabcolsep}{3pt}

\begin{adjustbox}{width=\textwidth}
\begin{tabular}{
  >{\raggedright\arraybackslash}m{2.1cm}!   
  *{4}{>{\centering\arraybackslash}m{0.7cm}}!   
  *{4}{>{\centering\arraybackslash}m{0.7cm}}!   
  *{4}{>{\centering\arraybackslash}m{0.7cm}}!   
  *{4}{>{\centering\arraybackslash}m{0.7cm}}!   
  *{4}{>{\centering\arraybackslash}m{0.7cm}}     
}
\toprule
\multirow{2}{*}{\textbf{Model}} &
\multicolumn{4}{c!}{\textbf{Overall}} &
\multicolumn{4}{c!}{\textbf{Object}} &
\multicolumn{4}{c!}{\textbf{Person}} &
\multicolumn{4}{c!}{\textbf{HOI}} &
\multicolumn{4}{c}{\textbf{De\&Re}} \\
\cmidrule(lr){2-5}\cmidrule(lr){6-9}\cmidrule(lr){10-13}\cmidrule(lr){14-17}\cmidrule(l){18-21}
& base & w/o & real & synth
& base & w/o & real & synth
& base & w/o & real & synth
& base & w/o & real & synth
& base & w/o & real & synth \\
\midrule
\multicolumn{21}{c}{\textbf{\textit{Open-Source Models}}} \\
\midrule
\rowcolor{red!22} BLIP3-o    
& 2.2 & 24.2 & \textbf{25.2} & 25.0       
& 3.2 & 40.3 & \textbf{40.3} & 40.2        
& 2.1 & 11.3 & \textbf{11.4} & 11.4       
& 2.1 & 23.9 & \textbf{25.0} & 24.9       
& 1.2 & 23.1 & \textbf{26.2} & 26.1 \\     

\rowcolor{red!22} DiMOO     
& 4.3 & 21.4 & \textbf{23.3} & 23.0   
& 3.8 & 38.4 & \textbf{38.4} & 38.2   
& 1.3 & 12.0 & \textbf{12.1} & 12.0   
& 6.5 & 24.1 & 24.7 & \textbf{24.8}  
& 1.8 & 18.2 & \textbf{21.3} & 21.0 \\ 

\rowcolor{orange!22} BAGEL     
& 19.3 & 32.7 & \textbf{34.4} & 34.2   
& 24.4 & 39.0 & \textbf{39.0} & 38.9   
& {22.4} & 28.3 & \textbf{28.5} & 28.3   
& {12.9} & 24.4 & \textbf{25.3} & 25.3   
& 34.2 & 41.7 & \textbf{44.5} & 44.2\\ 

\rowcolor{orange!22} Qwen-Image   
& {24.9} & 34.1 & \textbf{35.9} & 35.8   
& 39.4 & 52.3 & \textbf{52.4} & 52.3  
& {17.8} & 20.9 & \textbf{21.1} & 21.0  
& {20.0} & 34.6 & \textbf{35.0} & 34.9   
& {27.1} & 34.9 & \textbf{37.4} & 37.3 \\ 

\rowcolor{yellow!22} OmniGen2  
& {31.4} & 33.2 & \textbf{33.8} & 33.3  
& 44.2 & 46.2 & \textbf{46.3} & 46.2   
& {22.0} & 22.6 & \textbf{22.9} & 22.7   
& {27.4} & 31.8 & \textbf{32.2} & 32.2   
& {36.4} & 36.3 & \textbf{36.8} & 36.7 \\ 
\midrule
\multicolumn{21}{c}{\textbf{\textit{Closed-Source Models}}} \\
\midrule
\rowcolor{blue!22} GPT-4o        
& \textbf{47.1} &  --   &  --   &  --    
& \textbf{48.5} &  --   &  --   &  --    
& \textbf{36.2} &  --   &  --   &  --    
& \textbf{47.6} &  --   &  --   &  --    
& \textbf{50.3} &  --   &  --   &  --   \\
\rowcolor{blue!22} Nano-Banana   
& \textbf{47.8} &  --   &  --   &  --    
& \textbf{48.0} &  --   &  --   &  --    
& \textbf{41.8} &  --   &  --   &  -- 
& \textbf{49.6} &  --   &  --   &  --    
& \textbf{49.4} &  --   &  --   &  --   \\
\bottomrule
\end{tabular}
\end{adjustbox}
\end{table*}

\begin{figure*}[t!]
  \centering
  \includegraphics[width=0.999\textwidth]{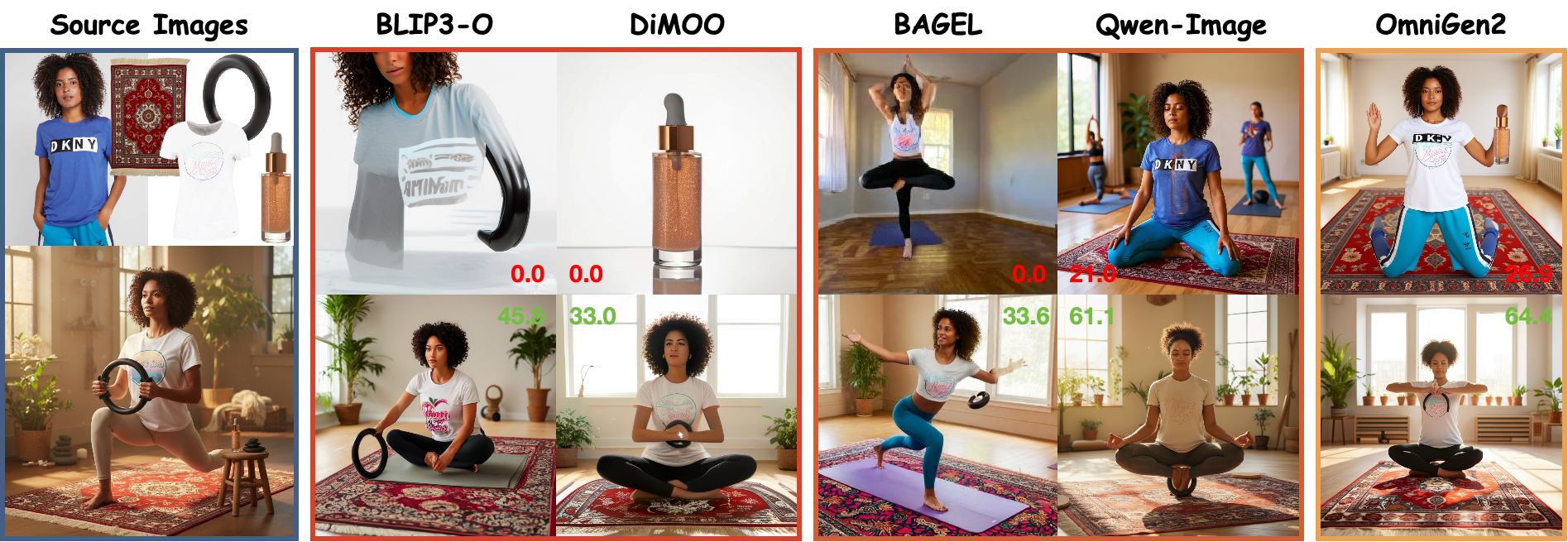}
  \vspace{-5mm}
  \caption{The leftmost displays the source and reference images. The first row shows model outputs before fine-tuning, the second row presents outputs after fine-tuning. The Weighted-Ref-VIEScore for each generated result is annotated in the corner. MICo-150K demonstrates strong robustness: \textbf{BLIP-3o} and \textbf{Lumina-DiMOO} acquire MICo capability from scratch; the emergent MICo abilities of \textbf{BAGEL} and \textbf{Qwen-Image} are significantly strengthened; \textbf{OmniGen2} achieves further improvement on top of its already strong performance.}
  \label{fig:train-case}
  \vspace{-3mm}
\end{figure*}

\section{Experiments}
\label{sec:Experiments}


\subsection{Experiments Setting}
To validate the effectiveness of our dataset, we train five open-source models: BAGEL~\citep{deng2025bagel}, OmniGen2~\citep{wu2025omnigen2explorationadvancedmultimodal}, Lumina-DiMOO~\citep{xin2025luminadimooomnidiffusionlarge}, BLIP3-o~\citep{chen2025blip3ofamilyfullyopen}, Qwen-Image-Edit~\citep{wu2025qwen-image} on MICo-150K, and evaluate their performance on the released MICo-Bench.
For BAGEL, we freeze the understanding branch and VAE while fine-tuning the generation branch parameters.
For BLIP3-o, we build upon the BLIP3o-Next-Edit~\citep{chen2025blip3onextfrontiernativeimage} version and set the maximum sequence length to 5120, with sequences right-padded or truncated as needed.
For OmniGen2, we simply perform full fine-tuning.
For Lumina-DiMOO, we set the input maximum sequence length to 6144, and resize all input images to $512 \times 512$.
For Qwen-Image-Edit, we use our custom training pipeline. The encoder and VAE are frozen, while all parameters of the MMDiT module are fine-tuned.
For comparison, we also evaluate representative closed-source systems and original open-source checkpoints under the same benchmark protocol.

\subsection{Results and Discussions}
The experimental results in Tab.~\ref{tab:mico_bench_results} show that closed-source systems still lead the released MICo-Bench, with Gemini-3-Pro-Image-Preview achieving the best overall score of 51.76.
Among open-source and MICo-finetuned models, our baseline \textbf{Qwen-Image-MICo} (derived from {Qwen-Image-Edit}~\citep{wu2025qwen-image}) achieves the highest overall score of 35.86 and substantially outperforms both Qwen-Image-Edit and Qwen-Image-2509 under the same evaluation protocol. 
Further quantitative experiments and qualitative examples of Qwen-MICo are provided in Sec.~\ref{app:Qwen-MICo-Evaluation}.


\vspace{+1mm}
\noindent\textbf{Open-Source Models}.
As shown in Fig.~\ref{fig:train-case},
{BLIP3-o} and {Lumina-DiMOO}, which originally lack any MICo ability, acquire usable multi-image composition behavior after fine-tuning on MICo-150K.
{BAGEL} and {Qwen-Image-Edit}, benefiting from large-scale multi-task pretraining, already display \textit{emergent MICo capabilities}, which are further reinforced after fine-tuning. 
The fine-tuned {Qwen-Image-MICo} obtains 35.86 overall, outperforming {Qwen-Image-2509}~\citep{qwen2025qwenimageedit2509} by 8.39 points and {Qwen-Image-Edit} by 10.92 points, despite Qwen-Image-2509 being trained on a much larger corpus. 
Moreover, Qwen-MICo supports arbitrary multi-image inputs, whereas Qwen-Image-2509 is limited to three.
{OmniGen2}, pretrained on MICo-related tasks, already exhibits compositional ability, especially in maintaining facial identity; its MICo-finetuned counterpart further improves the overall score from 31.42 to 33.82.

\vspace{+1mm}
\noindent\textbf{Closed-Source Models}.
Overall, Gemini-series image models and GPT-Image-1.5 form the top tier on MICo-Bench.
Gemini-3-Pro-Image-Preview ranks first overall and is strongest on Human-Centric and De\&Re cases, while GPT-Image-1.5 obtains the best Object-Centric score.
Representative closed-source cases are illustrated in Fig.~\ref{fig:gpt-vs-nano} of the Appendix.

\vspace{+1mm}
\noindent\textbf{Emergent Capability.}
It is worth noting that {BAGEL} and {Qwen-Image-Edit} were not trained on any multi-image composition data. However, \textit{when multiple source image tokens are simply concatenated and fed into the model, they exhibit emergent MICo ability}.
After simple SFT on MICo-150K, these abilities are significantly enhanced.
Moreover, even without using data from the De\&Re task, where no other training samples contain combined human, object, clothing, and scene all together, all models still demonstrate emergent performance on the MICo-Bench De\&Re subset.

\section{Conclusion}
\label{sec:Conclusion}
We introduced the \textbf{M}ulti-\textbf{I}mage \textbf{Co}mposition (\textbf{MICo}) task, along with \textbf{MICo-150K}, the first large-scale high-quality dataset tailored for this task. MICo-150K encompassed three tasks, seven subtasks, and 27 composition types, along with a \textbf{De\&Re} subset providing real–synthetic pairs for controlled comparison, totaling 150K MICo cases with varying numbers of input images.
We further proposed \textbf{MICo-Bench}, a 897-case benchmark, and \textbf{Weighted-Ref-VIEScore}, providing a reliable framework for evaluating MICo performance.
Extensive experiments revealed interesting insights and showed that finetuning on MICo-150K could consistently boost the MICo performance of different models with different initial capability.
Notably, our baseline model Qwen-MICo outperformed Qwen-Image-2509 on MICo-Bench while supporting arbitrary multi-image inputs beyond Qwen-Image-2509's three-image limitation.


\section{Appendix}
We organize this \textbf{Appendix} into the following sections:

\vspace{+1mm}
\begin{itemize}
    \item \textbf{\hyperref[app:Reproducibility]{A1. Reproducibility, Licensing, Data Release}}
    \item \textbf{\hyperref[app:Visualization]{A2. Visualizations and Qualitative Examples}}
    \item \textbf{\hyperref[app:Human-Face]{A3. Analysis of Human-Face Source Images}}
    \item \textbf{\hyperref[app:Weighted-Ref-VIEScore]{A4. Details of Weighted-Ref-VIEScore}}
    \item \textbf{\hyperref[app:humanstudy]{A5. Human Study and Metric Alignment}}
    \item \textbf{\hyperref[app:AdditionalExperiment]{A6. Additional Experiments}}
    \item \textbf{\hyperref[app:Qwen-MICo-Evaluation]{A7. Quantitative Evaluation of Qwen-MICo}}
\end{itemize}

\subsection{Reproducibility, Licensing, Data Release}
\label{app:Reproducibility}
MICo-150K and MICo-Bench (which are strictly non-overlapping—the
training set contains no MICo-Bench images) have been released for
research use. The release includes MICo-150K, the 897-case MICo-Bench annotations,
benchmark source/reference images, and the complete evaluation scripts and prompts.
All GPT-based evaluations in the released protocol are conducted with
\texttt{temperature=0}, yielding deterministic judge calls under a fixed evaluator model.

All MICo-150K source images come from publicly released datasets
with open licenses, including
Subject200k~\citep{tan2025ominicontrolminimaluniversalcontrol},
VITON-HD~\citep{choi2021vitonhdhighresolutionvirtualtryon},
X2I-Subject-Driven~\citep{xiao2024omnigenunifiedimagegeneration},
Mulan~\citep{tudosiu2024mulanmultilayerannotated},
Echo-4o~\citep{ye2025echo4o},
SUN397~\citep{5539970},
and CC12M~\citep{changpinyo2021conceptual12mpushingwebscale}.
The two headshot datasets used for identity filtering,
BKM1804/headshot\_istockphoto~\citep{bkm1804_headshot_istockphoto}
and
BKM1804/headshot\_pexels\_v1~\citep{bkm1804_headshot_pexels_v1},
are released under the \textbf{Creative Commons Attribution
4.0 (CC BY 4.0) license}. This license permits redistribution,
modification, and research use, provided that proper attribution is given.

Our use of these datasets conforms to all licensing and
redistribution requirements. No private or restricted data were used,
and no personally identifiable images outside openly licensed sources
were collected.

\subsection{Visualizations and Qualitative Examples}
\label{app:Visualization}
\begin{figure*}[t!]
  \centering
  \includegraphics[width=0.999\textwidth]{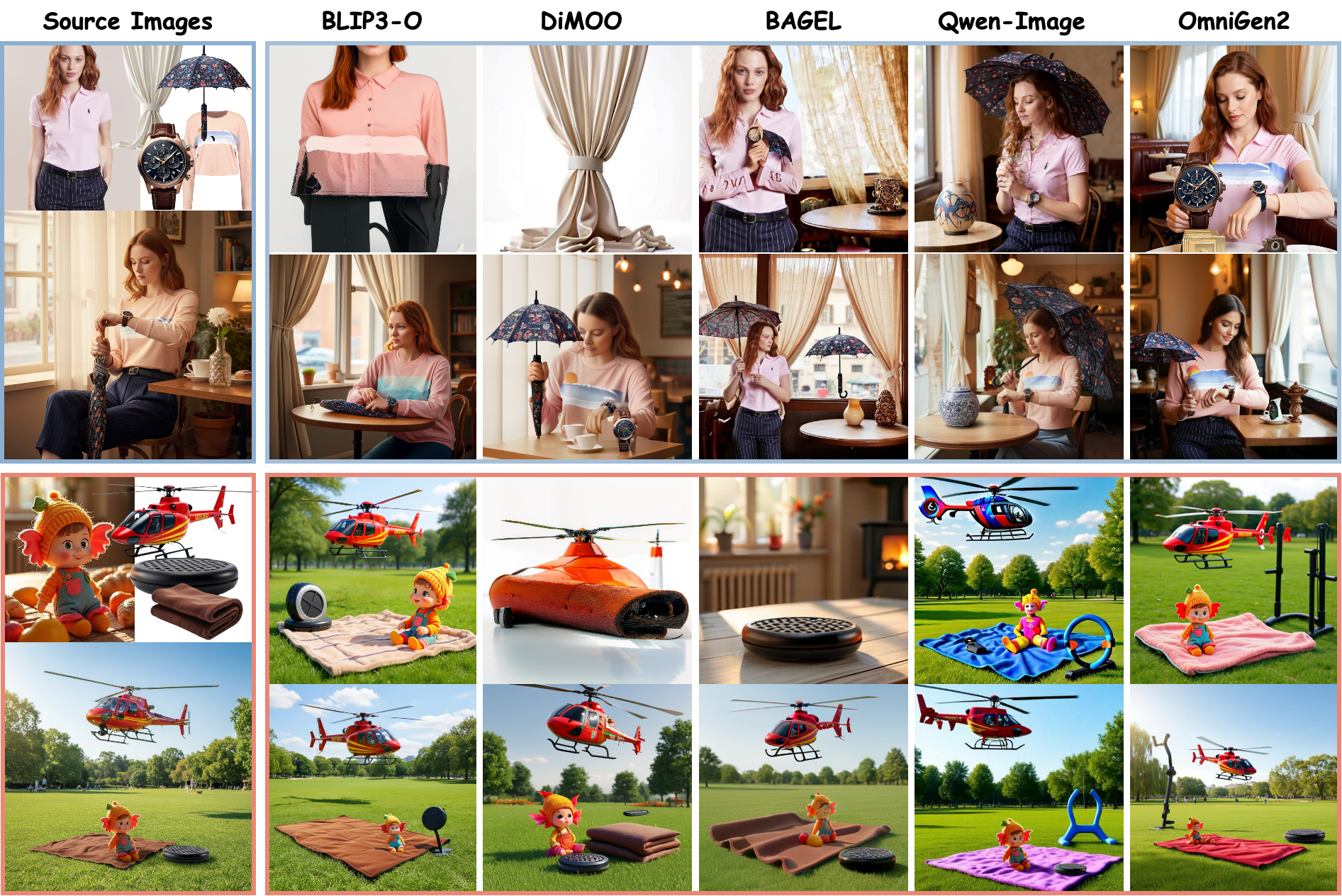}
  \caption{Comparison of open-source models before and after MICo-150K training. Some source images were cropped or background-removed for visualization. BLIP3-o~\citep{chen2025blip3ofamilyfullyopen} and Lumina-DiMOO~\citep{xin2025luminadimooomnidiffusionlarge} gain strong multi-image composition abilities after training. Qwen-Image-Edit~\citep{wu2025qwen-image} and BAGEL~\citep{deng2025bagel} were not explicitly trained for MICo tasks, but exhibit emergent MICo capabilities that are further enhanced through fine-tuning. OmniGen2~\citep{wu2025omnigen2explorationadvancedmultimodal} preserves identity well and produces more aesthetic, prompt-aligned results after training.}
  \label{fig:train-cases}
\end{figure*}

\begin{figure*}[t]
  \centering
  \includegraphics[width=0.999\textwidth]{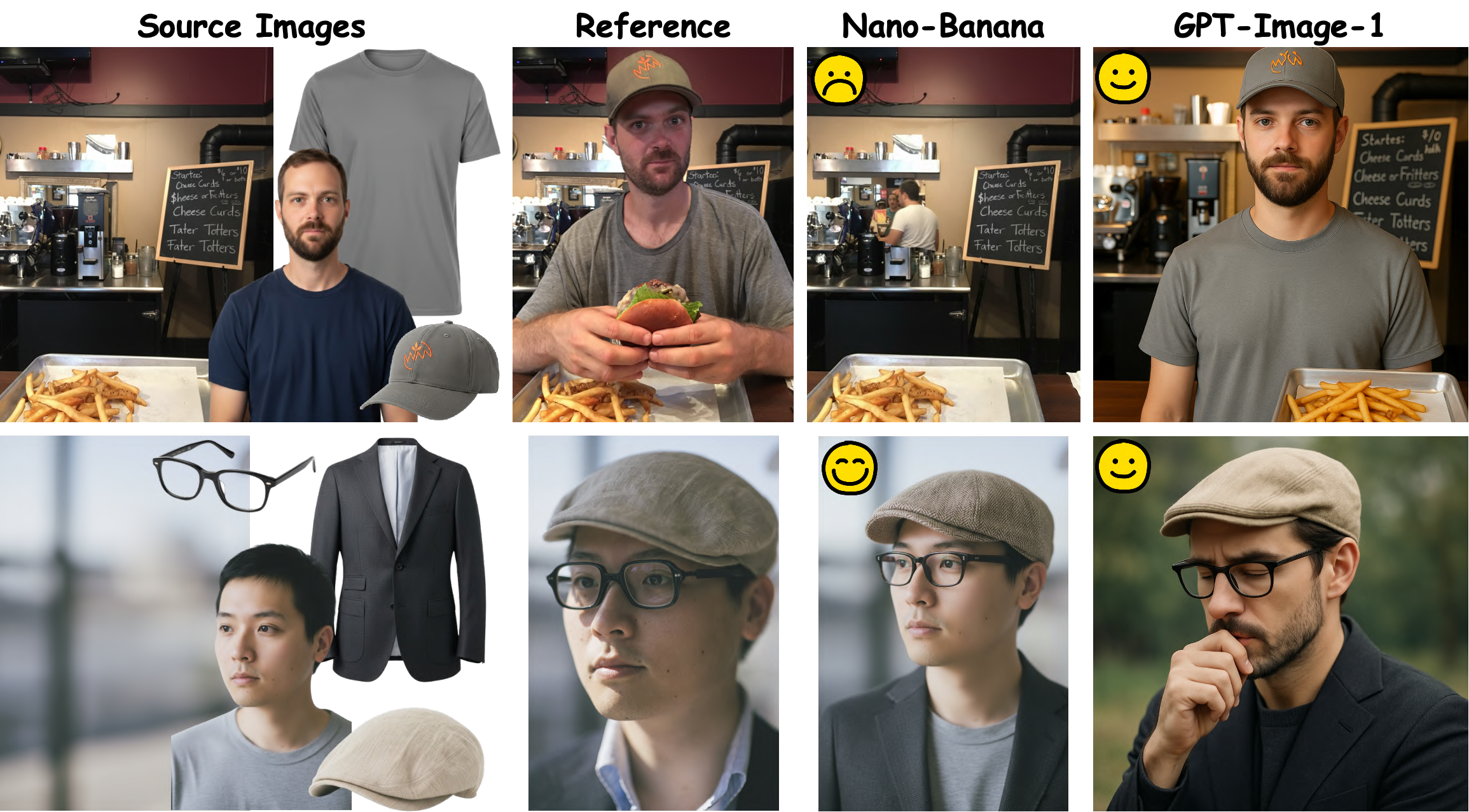}
  \caption{Nano-Banana~\citep{google2024nanobanana} produces more realistic images with stronger fidelity to the source inputs and a higher quality ceiling, but occasionally fails on certain cases. GPT-Image-1~\citep{openai2025gpt4o_image} exhibits a more stylized, less photo-realistic look, yet remains highly stable and consistently yields semantically coherent results.
}
  \label{fig:gpt-vs-nano}
\end{figure*}

We provide additional visual examples.
Fig.~\ref{fig:train-cases} illustrates examples from the five open-source models before and after being trained on MICo-150K.
Fig.~\ref{fig:gpt-vs-nano} compares examples generated by GPT-Image-1~\citep{openai2025gpt4o_image} and Nano-Banana~\citep{google2024nanobanana}.

\subsection{Analysis of Human-Face Source Images}
\label{app:Human-Face}
\subsubsection{Data Sources, Licensing, and Ethics}
\label{app:licensing}


\noindent\textbf{Celebrity Faces.}
All celebrity portraits in MICo\text{-}150K originate exclusively from the public X2I\text{-}Subject\text{-}Driven dataset~\citep{xiao2024omnigenunifiedimagegeneration}, 
which provides high\text{-}quality subject\text{-}centric photographs under a 
research\text{-}permissive license.  No additional celebrity images were crawled or collected from the web.
To mitigate right\text{-}of\text{-}publicity and privacy risks, 
\textbf{raw celebrity images will not be redistributed}; only derived metadata and 
selection indices will be released.

\vspace{+1mm}
\noindent\textbf{Non\text{-}Celebrity Faces.}
The remaining portraits are drawn from three established datasets commonly used in 
virtual try\text{-}on and portrait\text{-}generation research:
{VITON\text{-}HD}~\citep{choi2021vitonhdhighresolutionvirtualtryon} (research\text{-}only license),
{Headshot iStockPhoto}~\citep{bkm1804_headshot_istockphoto}
and {Headshot Pexels v1}~\citep{bkm1804_headshot_pexels_v1} (CC\text{-}BY\text{-}4.0).
The two headshot datasets released under \textbf{CC\text{-}BY\text{-}4.0} explicitly permit 
redistribution, modification, and research use with proper attribution, as 
discussed in Sec.~\ref{app:Reproducibility}. 
All images were used in strict accordance with their respective licenses.

Across all sources, MICo\text{-}150K contains 
\textbf{no private, scraped, or user\text{-}uploaded photographs}. 
The portraits are used solely for constructing multi\text{-}image composition tasks; 
no identity recognition, biometric profiling, or sensitive attribute inference is performed.

\subsubsection{Demographic Distribution Analysis}
\label{app:demographic-analysis}

To understand the demographic characteristics and potential bias of the human-face source images used in MICo-150K, we employed \textbf{Qwen2.5-VL-72B} to estimate three attributes for each portrait: (1) ethnicity, (2) gender, and (3) coarse age group. In total, we analyzed \textbf{53,648} human-face source images, including \textbf{11,677} images used in the De\&Re task and \textbf{41,971} images used in all other MICo tasks.

\vspace{1mm}
\noindent\textbf{Ethnicity.}
The estimated ethnic distribution is \textit{East/Southeast Asian: 6,740}, \textit{South Asian: 3,337}, \textit{European/Middle Eastern: 33,982}, \textit{African: 9,290} and \textit{Others: 299}. 
This reveals an over-representation of European/Middle Eastern subjects and relative under-representation of Asian groups, reflecting the typical skew of widely used portrait datasets.

\vspace{1mm}
\noindent\textbf{Gender.}
The raw gender distribution is \textit{Male: 21,241} and \textit{Female: 32,407}. During MICo-150K task construction, however, we {explicitly enforced gender balancing}. Across the {159,091} human-face images sampled for all compositions (see Tab.~\ref{tab:dataset_stats} in the main paper), the resulting ratio becomes {approximately 1:1}, mitigating upstream imbalance.

\vspace{1mm}
\noindent\textbf{Age.}
The age distribution is \textit{Child: 1,421}, \textit{Teen/Young Adult: 31,413}, \textit{Middle-Aged: 19,146}, and \textit{Older Adult: 1,668}, showing a strong dominance of young adults, a pattern that is common in fashion, try-on, and aesthetic portrait datasets. 
Importantly, for multi-image composition tasks, this age skew is {unlikely to negatively impact model behavior}, as real-world applications naturally exhibit a similar young-adult majority. Thus, while the distribution is uneven, it is largely aligned with practical usage scenarios.

\begin{figure}[H]
  \centering
  \includegraphics[width=\linewidth]{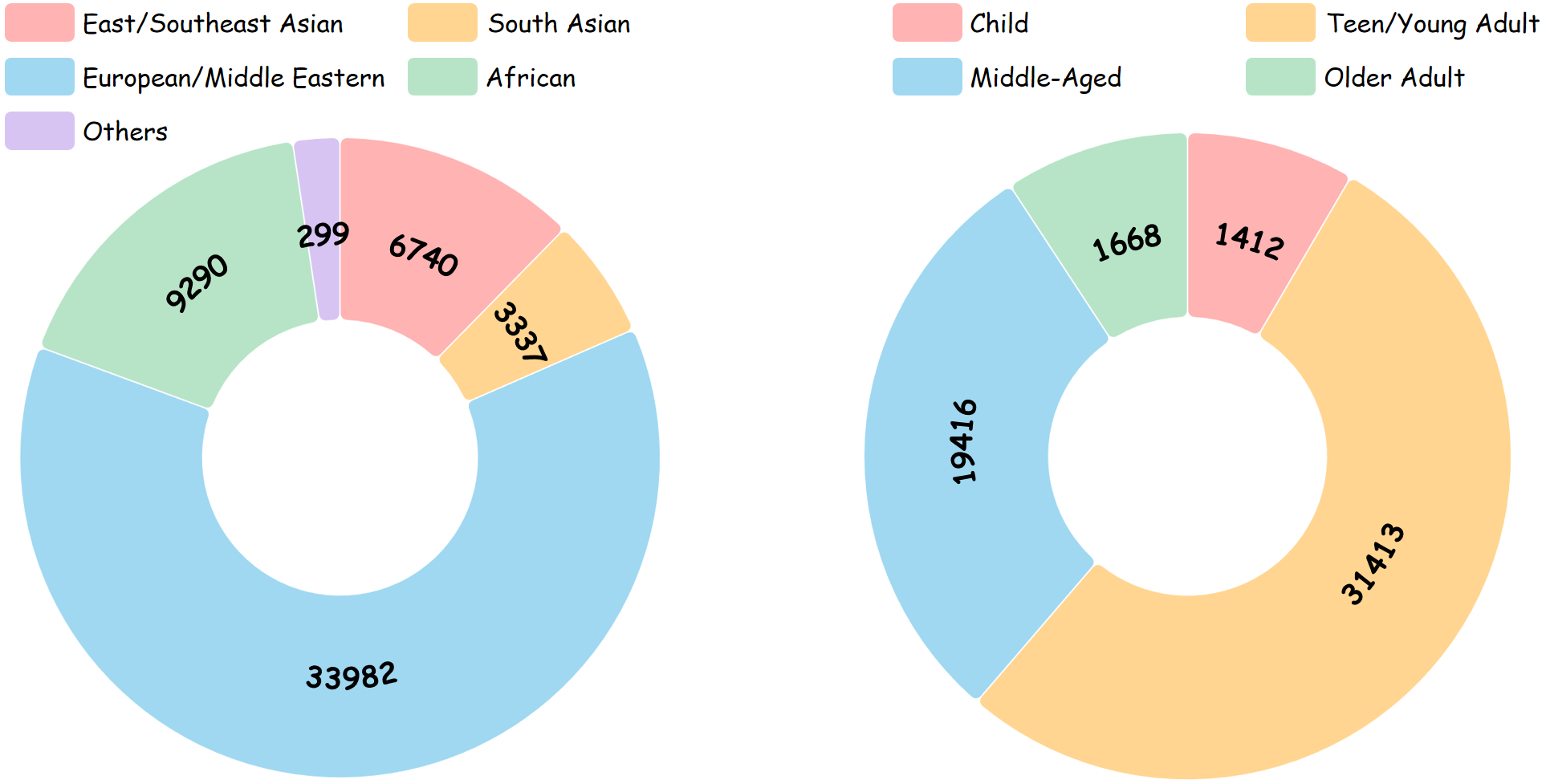}
    \caption{
Human-face source images exhibit a Western-centric ethnicity skew and a young-adult bias, reflecting characteristics inherent in the upstream public datasets rather than biases introduced by MICo-150K itself.
}
  \label{fig:human-face-analyze}
\end{figure}

\vspace{1mm}
\noindent\textbf{Discussion and Mitigation.}
The demographic analysis highlights several imbalances, particularly a Western-centric ethnicity skew and a young-adult bias. These limitations stem from upstream public datasets and are not specific to MICo-150K. We emphasize that MICo-150K is \textbf{not designed} for fairness benchmarking or identity-sensitive evaluations. Instead, its purpose is restricted to studying visual multi-image composition.

Overall, while demographic skew is present, as shown in Fig~\ref{fig:human-face-analyze}, we openly document these statistics to facilitate informed and responsible use. 
We caution against applying MICo-150K in fairness-critical or identity-sensitive contexts.
Future work may incorporate more diverse portrait sources to improve global representativeness.

\subsection{Details of Weighted-Ref-VIEScore}
\label{app:Weighted-Ref-VIEScore}

\subsubsection{Weights}
As described in Sec.~\ref{sec:WeightedRefVIEScore} of the main paper, the Weighted-Ref-VIEScore applies a multiplicative visibility coefficient \(W\), computed as the average preservation score over all source elements.
The critical step is therefore determining whether each source element is present and properly preserved in the generated target image.

\vspace{2mm}
\noindent\textbf{Objects, clothing items, and scenes.}
For non-human content, we compare each source image with the generated target image using a VLM.
In the released evaluation script, the default VLM is Qwen3-VL-30B-A3B-Instruct, while other strong VLMs can be substituted through the same yes/no interface.
For each source element, we ask whether the element from the source image appears in the generated target image and treat the model's answer as a binary indicator of presence.

\vspace{2mm}
\noindent\textbf{Human faces.}
For source images containing faces, we use {ArcFace}~\citep{Deng_2022} to extract
identity embeddings from both source and target portraits and compute
cosine similarity.
Rather than binarizing face preservation with a single threshold, we map the best matched ArcFace cosine similarity to a graded score:
\([0.45,1.0)\rightarrow1.0\), \([0.30,0.45)\rightarrow0.7\), \([0.15,0.30)\rightarrow0.5\), \([0.05,0.15)\rightarrow0.2\), and values below \(0.05\rightarrow0.0\).
The final weight is
\[
W=\frac{1}{N}\sum_{i=1}^{N}s_i,
\]
where \(N\) is the number of source images and \(s_i\) is either a graded face-preservation score or a binary non-face preservation score.

\subsubsection{SC and PQ Scoring}
To compute the {Semantic Consistency (\textbf{SC})} and {Perceptual Quality (\textbf{PQ})} components of the Weighted-Ref-VIEScore, we query GPT-5.4 twice, once for SC and once for PQ.  

For SC scoring, although current general-purpose VLMs~\citep{zhang2024mavis, lin2024draw, Wang_2024_CVPR, lin2025perceive, tong2025delving, wang2024mr, Lin_2023_ICCV, an2025unictokens, an2024mc, luo2024llm} are very powerful, they still struggle to perform reliable cross-image attention. Therefore, we use only a single reference image.
Specifically, the model takes two images as input, a reference image and the image to be evaluated, and outputs two sub-scores:

\begin{itemize}
    \item \textbf{Prompt Following (PF):} how well the generated image follows the textual input prompt;
    \item \textbf{Subject Resemblance (SR):} how well the generated image matches the reference image in terms of people, scenes, clothing, and objects.
\end{itemize}

Both PF and SR are scored on a \([0, 10]\) scale, and the final SC score is defined as $\text{SC} = \sqrt{\text{SR} \times \text{PF}}$.
We use a fixed meta-prompt for SC evaluation. The full prompt template is provided in our fully open-source \href{https://github.com/A113N-W3I/MICo-150K}{\textcolor{blue}{repo}}.

For PQ scoring, the model takes only the image to be evaluated as input and produces two sub-scores:

\begin{itemize}
\item \textbf{Naturalness:} evaluates whether the generated image appears visually plausible, including coherent lighting, shadows, geometry, and overall realism.
\item \textbf{Artifacts:} measures the absence of visual defects such as distortions, duplicated limbs, or unnatural textures.
\end{itemize}

\begin{figure}[H]
  \centering
  \includegraphics[width=\linewidth]{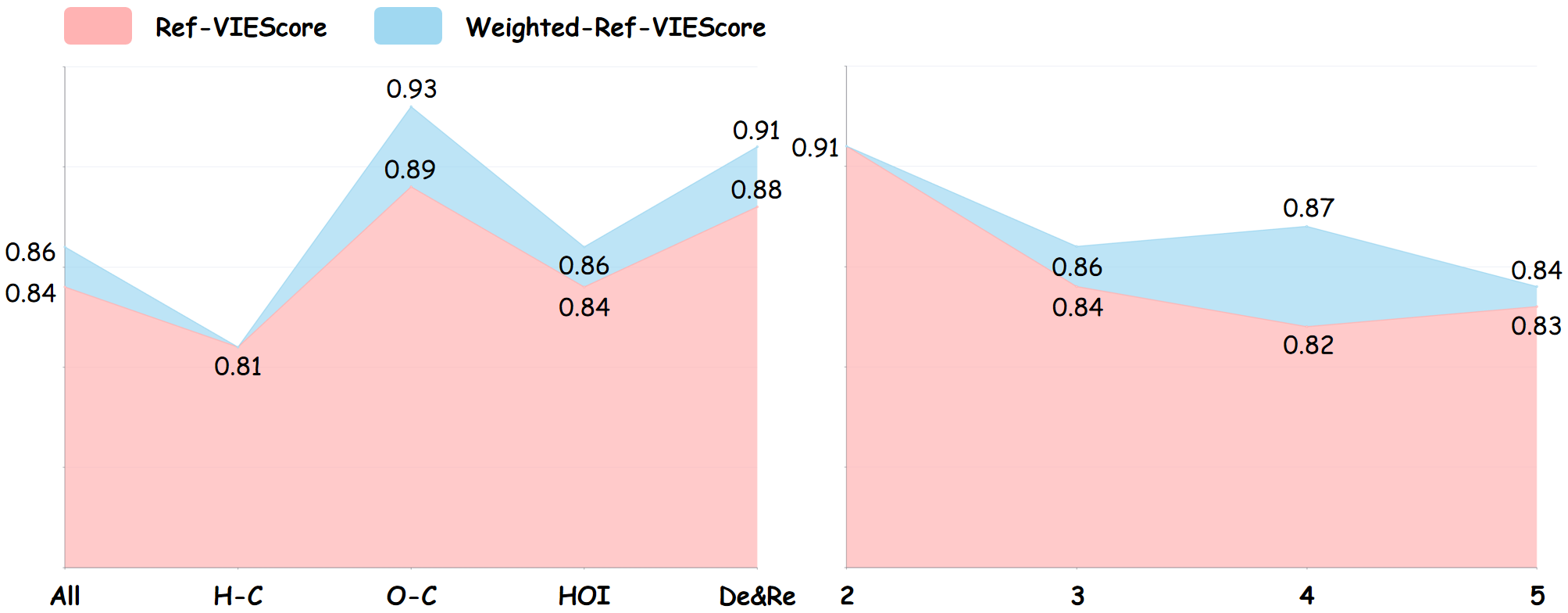}
    \caption{
We refer to the metric without the weighting factor $W$ as \textbf{Ref-VIEScore}. We conduct extensive human studies and compute the \textit{Spearman rank correlation} between human preferences and the rankings produced by both \textbf{Weighted-Ref-VIEScore} and \textbf{Ref-VIEScore}. The results are summarized in the figure.
}
  \label{fig:spearman}

\end{figure}

Both naturalness and artifacts are scored on a [0,10] scale, and the final PQ score is defined as their geometric mean.
The final per-case score is \(\text{Score}=W\times\text{SC}\times\text{PQ}\), which lies on a 0--100 scale.
The full prompt template is in our open-source \href{https://github.com/A113N-W3I/MICo-150K}{\textcolor{blue}{repo}}.

\subsection{Human Study and Metric Alignment}
\label{app:humanstudy}
To validate the effectiveness of {Weighted-Ref-VIEScore}, we conducted an extensive human study to measure its alignment with human preferences. We sampled {75 evaluation cases} covering the fine-grained MICo task types and additional De\&Re cases.

For each case, we randomly selected {five candidate outputs} from a pool of {twelve models}: five open-source models (BLIP-3o~\citep{chen2025blip3ofamilyfullyopen}, Lumina-DiMOO~\citep{xin2025luminadimooomnidiffusionlarge}, BAGEL~\citep{deng2025bagel}, Qwen-Image-Edit~\citep{wu2025qwen-image}, OmniGen2~\citep{wu2025omnigen2explorationadvancedmultimodal}) in both their pre-training and MICo-finetuned versions, and two closed-source models (GPT-Image-1~\citep{openai2025gpt4o_image} and Nano-Banana~\citep{google2024nanobanana}). Human evaluators were provided with the \textit{source images}, \textit{text prompts}, and the \textit{five anonymized candidate outputs} for ranking. 

We recruited {25 human participants}, all holding at least a bachelor's degree, including {9 senior Ph.D.\ students}. We further define {Ref-VIEScore} as the variant of our metric \textit{without} the weighting factor $W$, \ie, $\text{SC} \times \text{PQ}$.

For each case, we computed the \textbf{Spearman rank correlation} between each participant's ranking and the rankings produced by Weighted-Ref-VIEScore and Ref-VIEScore. 
The {average correlation across 25 participants} was used as the final correlation score for that case. We report results under two groupings: 
(1) grouped by \textbf{task type}, and 
(2) grouped by the \textbf{number of input source images}.
The results, summarized in Fig.~\ref{fig:spearman}, reveal three key findings:

\begin{enumerate}[leftmargin=*]
    \item \textbf{Strong human--metric alignment.}  
    Weighted-Ref-VIEScore achieves consistently high Spearman correlations across all task types, indicating robust agreement with human judgment.

    \item \textbf{Robustness to varying numbers of source images.}  
    Thanks to our reference-image design, Weighted-Ref-VIEScore maintains stable accuracy regardless of how many source images are provided as input.

    \item \textbf{Importance of the weighting factor $W$.}  
    Weighted-Ref-VIEScore outperforms Ref-VIEScore across all settings.  
    Moreover, the inclusion of $W$ provides interpretability by revealing which source elements failed to appear in the generated image, a capability essential for diagnosing and improving model behavior.
\end{enumerate}

\begin{table*}[t]
  \centering
  \caption{
We evaluate on the MICo-Bench subset where each case contains exactly three input images, since Qwen-Image-2509 does not support higher-order composition. Qwen-MICo consistently outperforms Qwen-Image-2509 across nearly all evaluation dimensions.
}
  \label{tab:vs2509_results}
  \newcolumntype{L}[1]{>{\raggedright\arraybackslash}p{#1}}
\newcolumntype{C}[1]{>{\centering\arraybackslash}p{#1}}
\newcommand{\catcell}[2]{\cellcolor{#1!22}\textbf{#2}}
\newcommand{\tripleshade}[4]{ & \cellcolor{#1} #2 & \cellcolor{#1} #3 & \cellcolor{#1} #4 \\}

\renewcommand{\arraystretch}{1.0}
\setlength{\extrarowheight}{-1pt}
\setlength{\tabcolsep}{5pt}
\renewcommand{\arraystretch}{1.0}

\adjustbox{max width=\linewidth}{%
\begin{tabular}{
    L{2.9cm}
    C{1.4cm} C{1.4cm} |            
    C{0.9cm} C{0.9cm} C{0.9cm} C{0.9cm} C{0.9cm} |  
    C{0.9cm} C{0.9cm} C{0.9cm} C{0.9cm}             
}
\toprule
\multirow{2}{*}{\textbf{Method}} &
\multicolumn{2}{>{}C{2.8cm}}{\hspace{0.5cm}\textbf{Object Centric}} &
\multicolumn{5}{>{}C{4.5cm}}{\hspace{1.0cm}\textbf{Person Centric}} &
\multicolumn{4}{>{}C{3.6cm}}{\hspace{0.8cm}\textbf{HOI}} \\
&
 2O1S &  3O &
 2M1W &  2W1M &
 3M   &  3W   &  2P1S &
 1P2O &  2P1O &
 1P2C &  1P1C1O \\
\midrule
Qwen-MICo       &
\textbf{56.12} & \textbf{59.56} &
\textbf{59.04} & \textbf{58.96} & \textbf{50.11} & \textbf{56.19} & \textbf{60.97} &
\textbf{54.92} & 52.16 & \textbf{55.82} & \textbf{54.26} \\
Qwen-Image-2509 &
56.00 & 45.32 &
42.63 & 52.46 & 48.40 & 50.78 & 49.70 &
50.64 & \textbf{54.65} & 51.77 & 47.91 \\
\bottomrule
\end{tabular}%
}

\end{table*}

\subsection{Additional Experiments}
\label{app:AdditionalExperiment}
\subsubsection{Reference Image and Style Homology Risk}
\label{app:Model Style Homology Risk}

A natural concern is whether Weighted-Ref-VIEScore may implicitly favor images whose visual style is closer to that of the reference generator, since reference images for most MICo-Bench tasks are synthesized using Nano-Banana~\citep{google2024nanobanana}. 
One might hypothesize that models producing Nano-Banana–like aesthetics could receive higher scores.

We first clarify the setup: for all De\&Re task cases, the reference images are real photographs, whereas only the reference images for the remaining 27 task types are generated by Nano-Banana.

To evaluate whether such style homology affects scoring, we randomly sampled \textbf{108 cases} from the non-De\&Re portion of MICo-Bench. 
For each case, we generated outputs from three different models, {GPT-Image-1~\citep{openai2025gpt4o_image}, Qwen-MICo, and Nano-Banana~\citep{google2024nanobanana}}. 
We then computed a {style distance} between each model output and the corresponding reference image using {DINOv2~\citep{oquab2024dinov2learningrobustvisual} ViT-b} cosine distance, a strong self-supervised perceptual feature encoder widely used for measuring cross-image similarity.

This produced 
a sequence of model rankings induced by style distance and 
a sequence of rankings induced by Weighted-Ref-VIEScore. 
We computed the {Spearman correlation} $\rho$ between these two rankings for each case.

\begin{figure*}[h]
  \centering
  \includegraphics[width=0.999\textwidth]{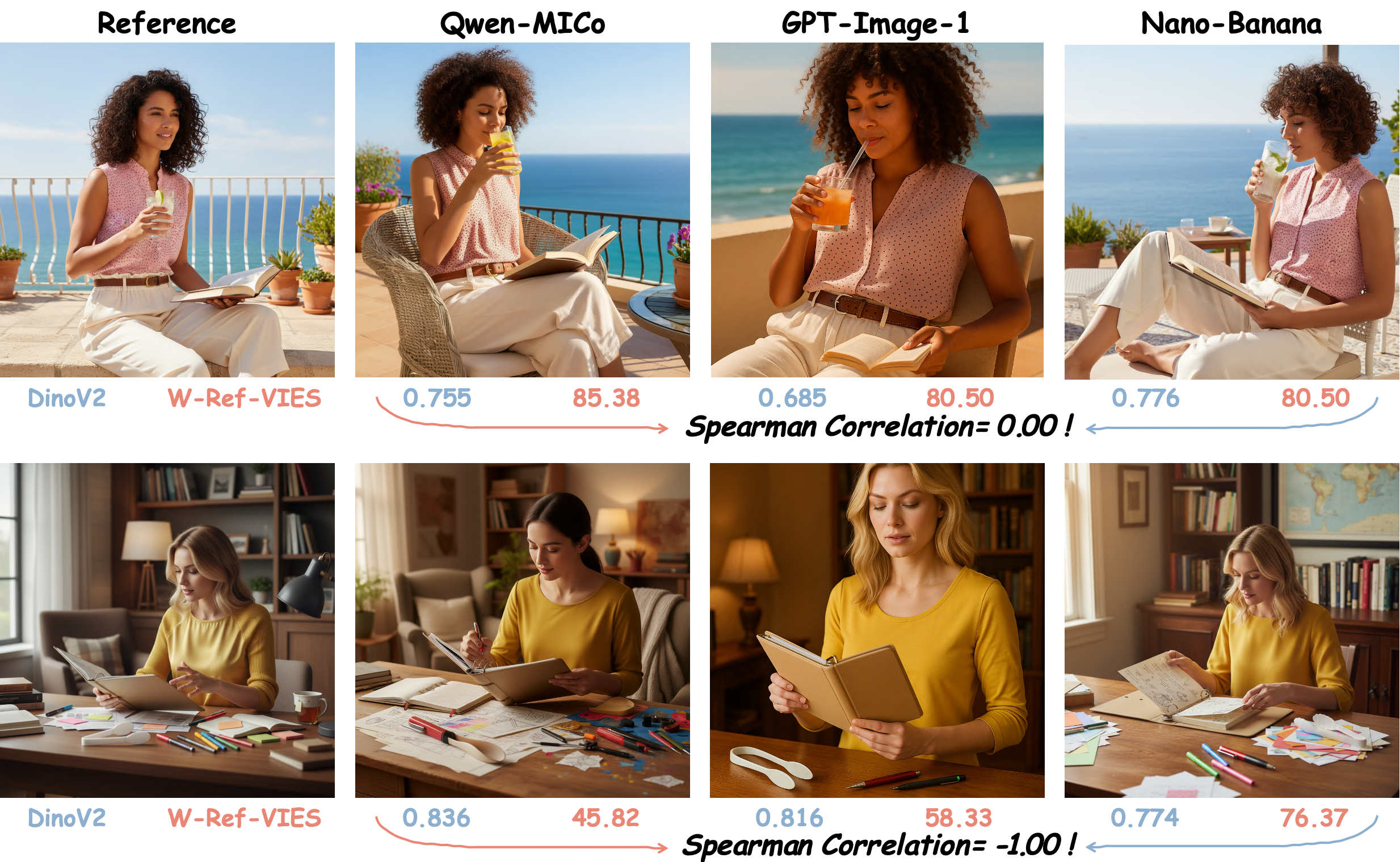}
  \caption{Similarity to the reference image does not influence how Weighted-Ref-VIEScore ranks model outputs, demonstrating that the metric remains objective and well-aligned with human preferences.
}
  \label{fig:ref-spearman}
\end{figure*}

Across all 108 cases, the mean correlation was \textbf{$\rho = 0.251$}, indicating only a weak association between reference-style similarity and VIEScore ranking. In other words, models whose outputs resemble the Nano-Banana reference style are \textbf{not} systematically favored. Weighted-Ref-VIEScore does \textbf{not} reward stylistic imitation, and its scoring behavior is largely orthogonal to reference similarity. Qualitative examples illustrating this lack of dependency are shown in Fig.~\ref{fig:ref-spearman}.

\subsubsection{VLM Evaluator-Generator Coupling}
Since our evaluation relies on a VLM judge to compute both the \textbf{PQ} (Perceptual Quality) and \textbf{SC} (Semantic Consistency) scores, a natural concern is the potential risk of \emph{evaluator–generator coupling}: 
if a generative model produces images whose style is similar to that of GPT-Image-1~\citep{openai2025gpt4o_image} (or when GPT-Image-1 itself is being evaluated), could such outputs receive artificially inflated scores?
Although prior works~\citep{wei2025tiif, hu2024ellaequipdiffusionmodels, wang2025videoverse, guo2025video} suggest that GPT-based evaluators do \textit{not} inherently favor GPT-generated content, we explicitly examine this possibility in the MICo evaluation setting.

Following the procedure in Sec.~\ref{app:Model Style Homology Risk}, we sampled the same 108 non-De\&Re cases from MICo-Bench and obtained outputs from three models: GPT-Image-1~\citep{openai2025gpt4o_image}, Qwen-MICo, and Nano-Banana~\citep{google2024nanobanana}. For each case, we computed a {style distance} by measuring the DINOv2 feature distance between each model’s output and the output generated by GPT-Image-1 for the same case.

We then computed the correlation between this style-distance sequence and three evaluation-score sequences:
SC scores, PQ scores, and Ref-VIEScore (SC $\times$ PQ). The resulting Spearman correlations were: $\rho_{\text{SC}} = -0.115,\quad \rho_{\text{PQ}} = -0.008,\quad\rho_{\text{Ref-VIEScore}} = -0.077$.

\begin{figure*}[h]
  \centering
  \includegraphics[width=0.999\textwidth]{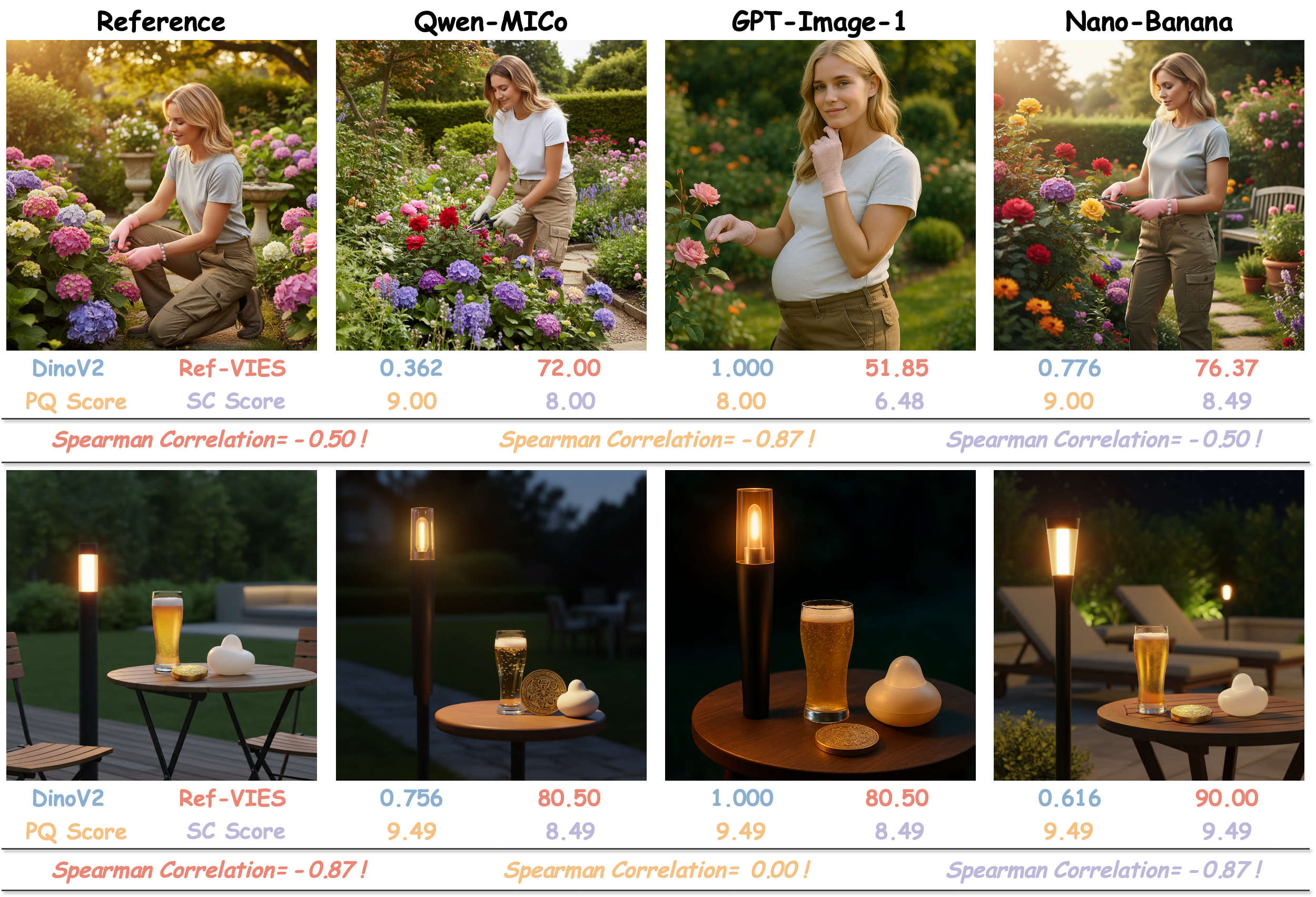}
  \caption{Although PQ and SC scores are computed using a GPT-based VLM judge, we find no evidence of evaluator–generator coupling: images that are stylistically closer to GPT-Image-1 do \emph{not} receive higher scores. The evaluation depends solely on MICo output quality and remains well aligned with human preferences.
}

  \label{fig:ref-gpt}
  \vspace{-2mm}
\end{figure*}

All correlations are near zero, indicating that neither SC, PQ, nor Ref-VIEScore favors outputs stylistically closer to GPT-generated images, even though a GPT-based VLM is used as the evaluator. 
This confirms the absence of evaluator-generator coupling in the MICo evaluation setting. Representative qualitative examples are shown in Fig.~\ref{fig:ref-gpt}.

\subsubsection{Resistance to Copy-Paste Hack}
A common failure mode for composition models is the \emph{copy-paste hack}: 
directly cutting objects or faces from the source images and pasting them onto a new background.  An effective metric must not reward such behavior.

To test whether Weighted-Ref-VIEScore can be fooled by this shortcut, we constructed explicit copy-paste baselines.  
For two representative sub-tasks, \textbf{Object+Scene} and \textbf{Person+Scene}, we randomly sampled 10 cases each.  
For every case, we segmented the source objects (or person) using an off-the-shelf segmentation tool~\citep{ravi2024sam2segmentimages} and manually pasted them onto the scene image, creating a naive composite without any harmonization.

\begin{figure*}[h]
  \centering
  \includegraphics[width=0.999\textwidth]{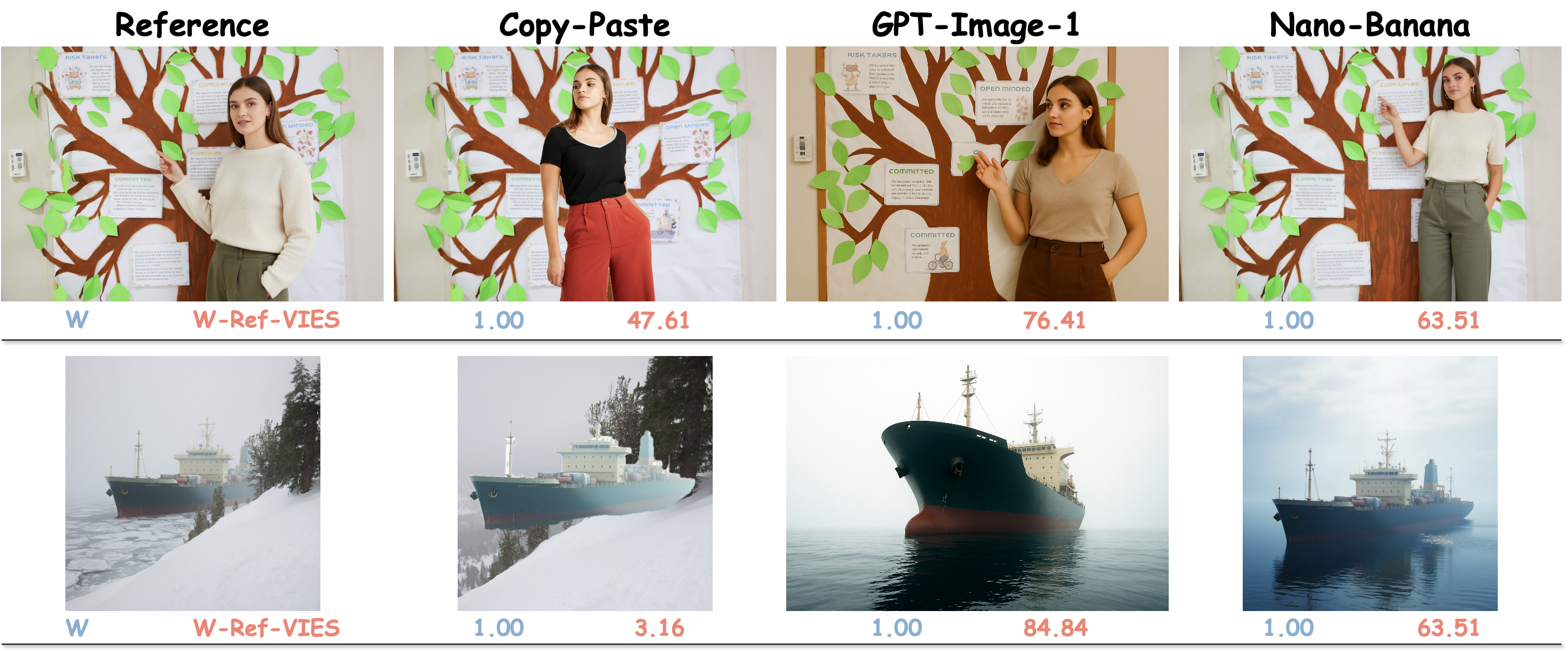}
  \caption{Weighted-Ref-VIEScore effectively \textbf{prevents copy--paste hacks}.  
We segmented objects or persons from the source images and manually pasted them onto the scene image to form a naïve, unharmonized composite.  
Although such copy--paste results achieve a perfect weight factor (since every source element appears in the output), their PQ and SC scores remain very low.
}

  \label{fig:copy-paste}
\end{figure*}

Although these copy-paste outputs achieve a perfect weight factor $W = 1.00$ (since all source elements appear in the target), their \textbf{PQ} and \textbf{SC} scores are extremely low.  
Across all 20 constructed cases, the final Weighted-Ref-VIEScore averages only \textbf{14.16}, demonstrating that the metric effectively penalizes unnatural compositing and cannot be exploited by trivial cut-and-paste strategies. Representative examples are shown in Fig.~\ref{fig:copy-paste}.

\subsection{Quantitative Evaluation of Qwen-MICo}
\label{app:Qwen-MICo-Evaluation}
For fair comparison, we clarify that Qwen-Image-2509~\citep{qwen2025qwenimageedit2509} was released before MICo-150K became available, and thus it could not have used our dataset for training. This eliminates the possibility of data leakage or overlap between MICo-150K and Qwen-Image-2509's training corpus.

To further quantify the performance gap between Qwen-Image-2509~\citep{qwen2025qwenimageedit2509} and Qwen-MICo, we evaluate both models on the subset of MICo-Bench containing tasks with exactly three input images (see Tab.~\ref{tab:dataset_stats} in the main paper). This subset includes 11 task types, \textit{2O1S, 3O, 2M1W, 2W1M, 3M, 3W, 2P1S, 1P2O, 2P1O, 1P2C, 1P1C1O}. For each type, we randomly sample five cases, yielding a total of 55 evaluation instances. We then compute the proposed Weighted-Ref-VIEScore for both models across all cases.

\begin{figure*}[h]
  \centering
  \includegraphics[width=0.999\textwidth]{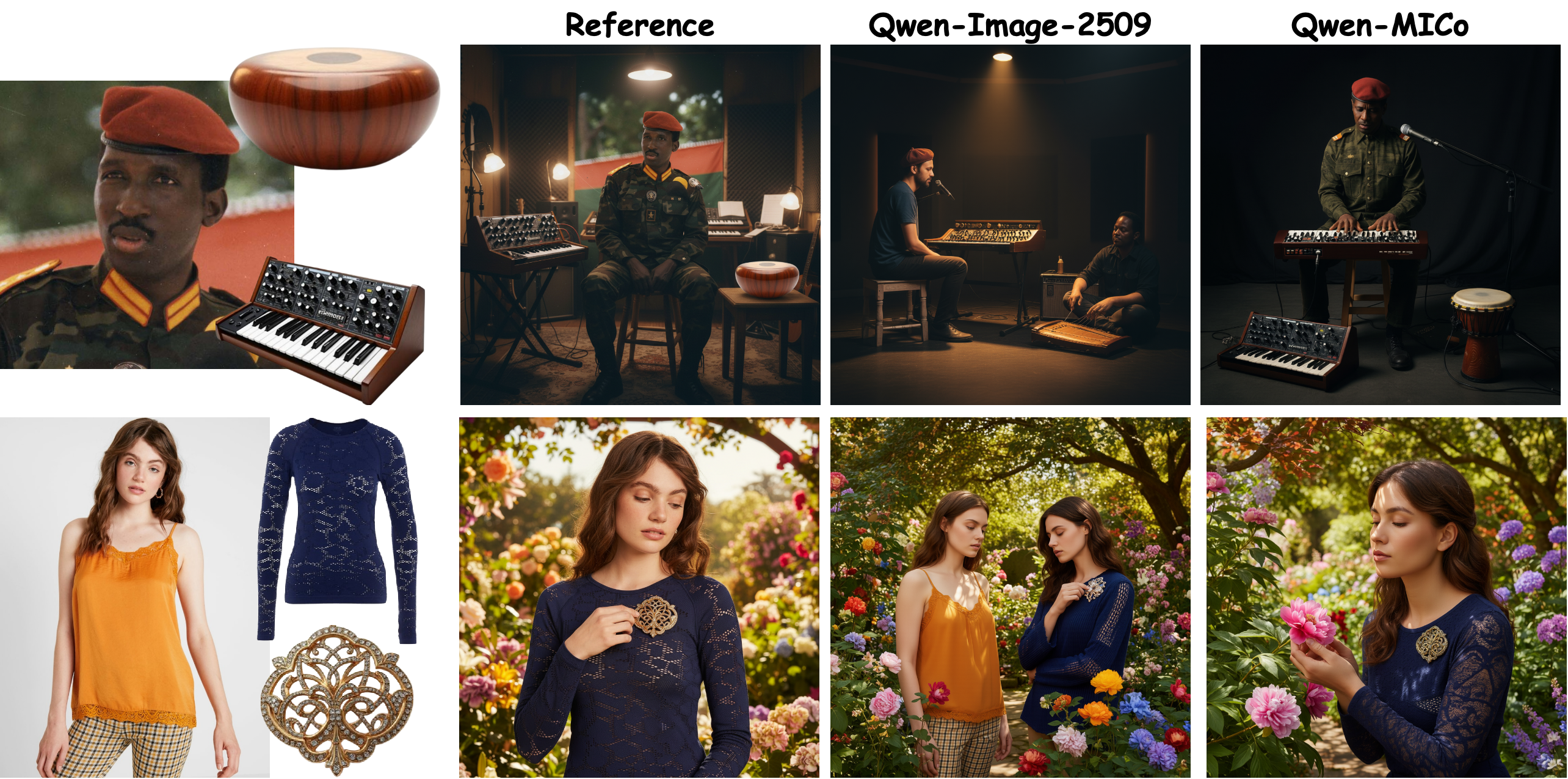}
  \caption{Qwen-MICo consistently outperforms Qwen-Image-2509 across nearly all evaluation dimensions on the MICo-Bench three-image subset. While Qwen-Image-2509 is trained on a massive corpus but restricted to three-image inputs, Qwen-MICo—trained only on MICo-150K—supports arbitrary multi-image composition and yields higher compositional fidelity and visual quality.
}

  \label{fig:appvs2509}
\end{figure*}

As reported in Tab~\ref{tab:vs2509_results}, Qwen-MICo consistently outperforms Qwen-Image-2509 across nearly all evaluation dimensions, despite using two to three orders of magnitude less training data and supporting arbitrary numbers of input images, whereas Qwen-Image-2509 is limited to three-image composition.
Additional qualitative comparisons are shown in Fig.~\ref{fig:appvs2509}, illustrating the superior compositional fidelity and visual coherence achieved by Qwen-MICo.

\begin{figure*}[t]
  \centering
  \includegraphics[width=\linewidth]{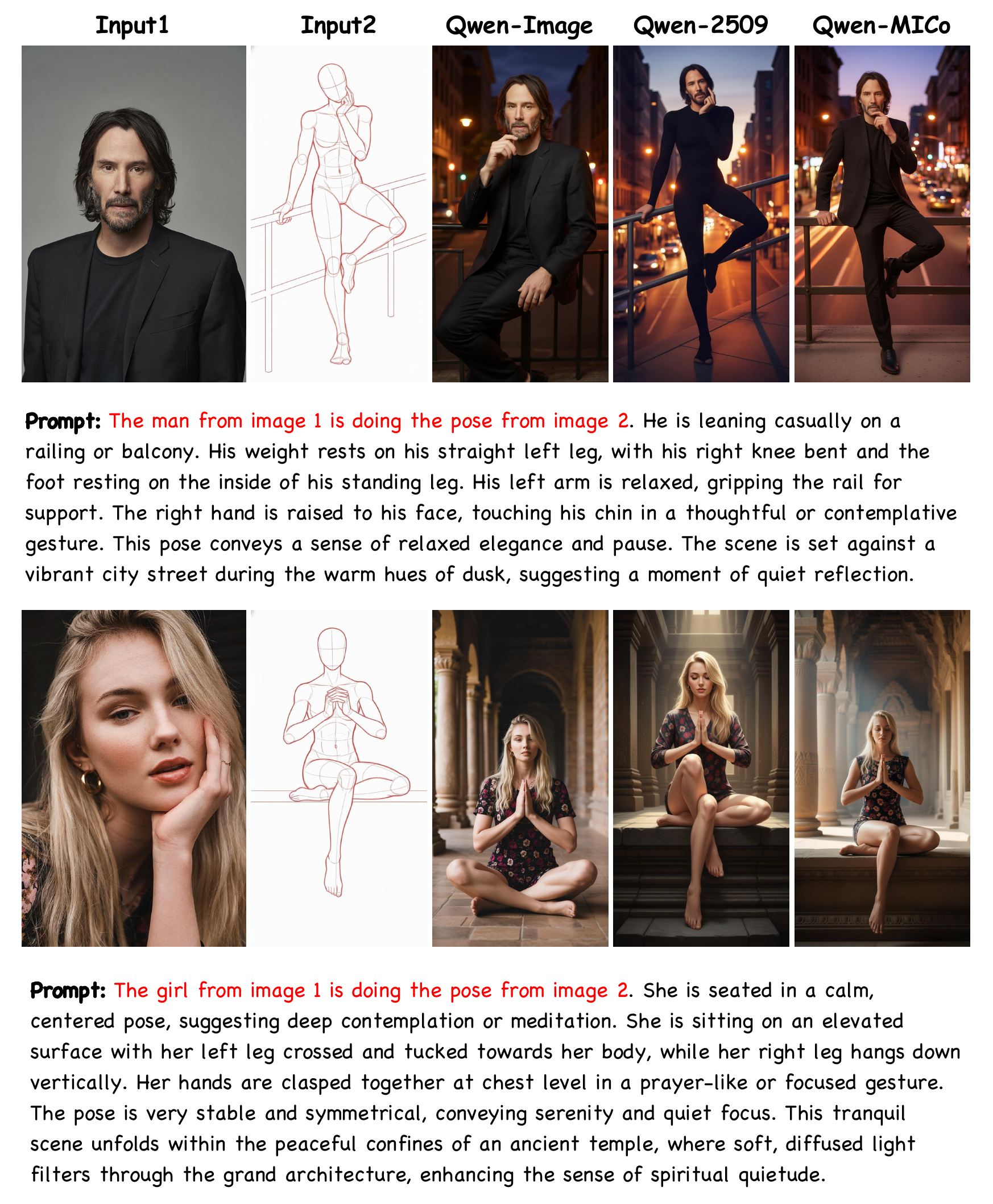}
  \caption{Qwen-MICo exhibits strong emergent abilities in recognizing and composing complex \textbf{human poses}.}

  \label{fig:mico-case-1}
\end{figure*}

\begin{figure*}[t]
  \centering
  \includegraphics[width=\linewidth]{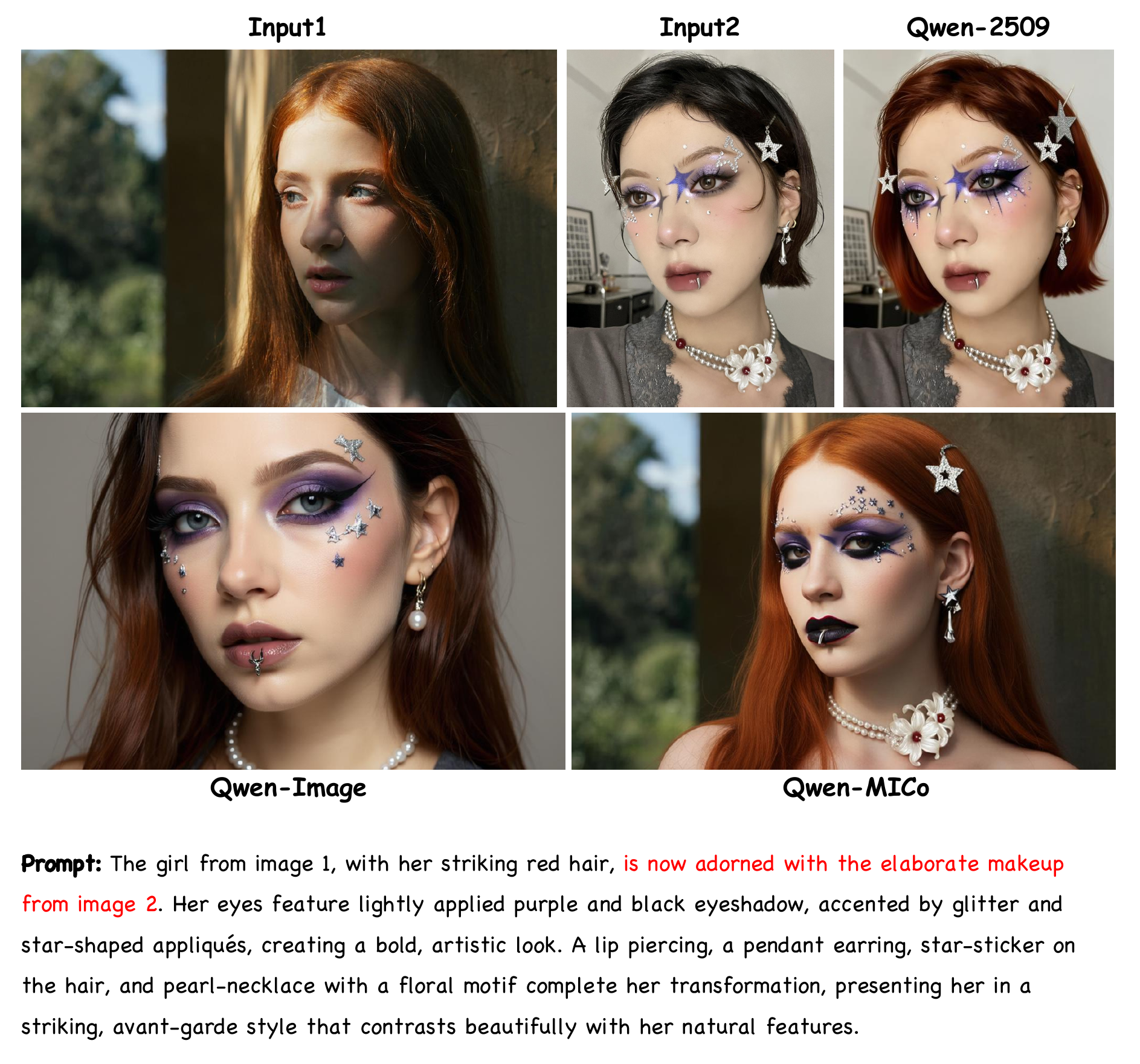}
  \caption{Qwen-MICo performs well on \textbf{virtual makeup try-on} (transferring the makeup in Image~2 onto the girl in Image~1).}

  \label{fig:mico-case-2}
\end{figure*}

\begin{figure*}[t]
  \centering
  \includegraphics[width=\linewidth]{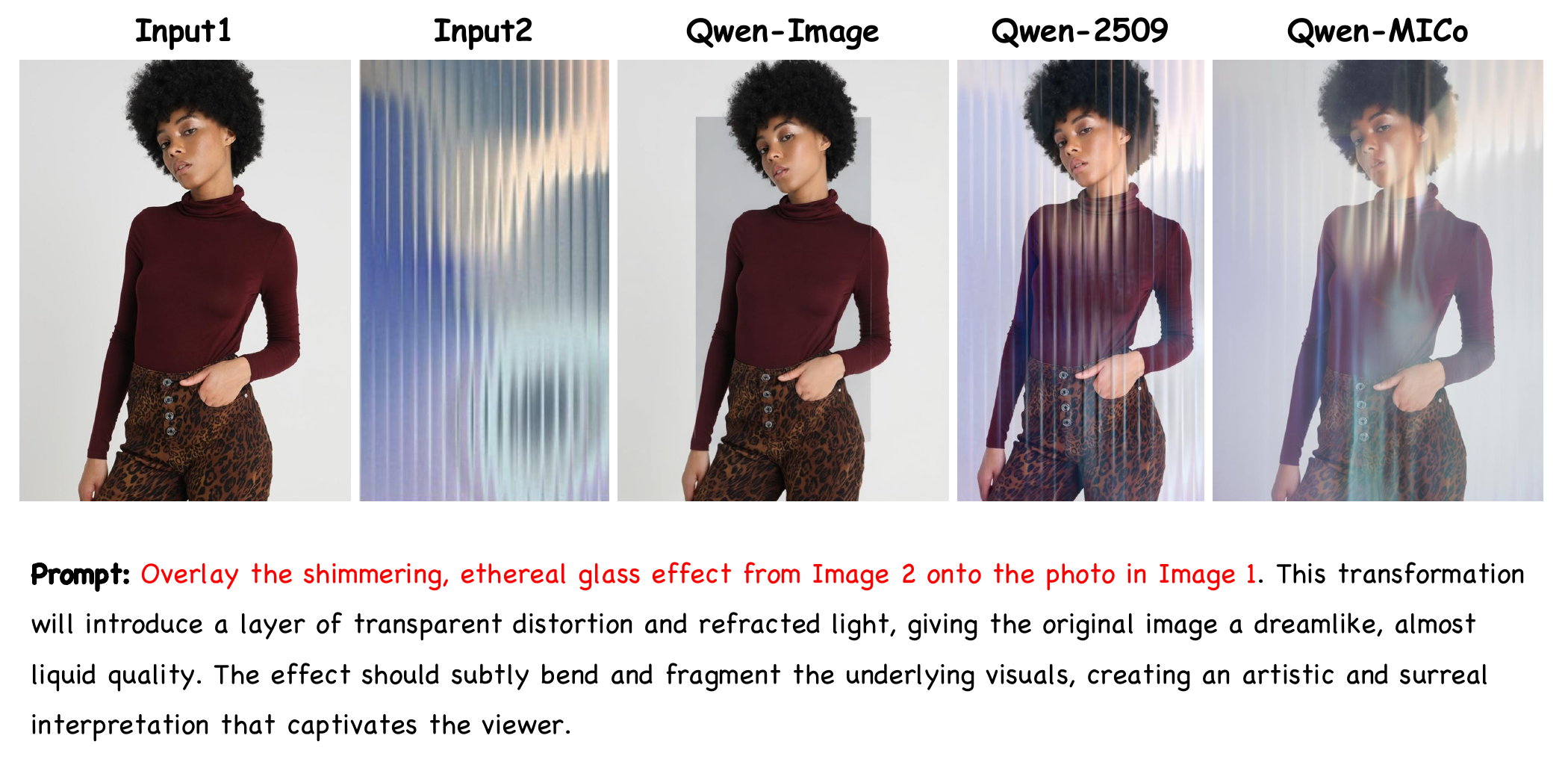}
  \caption{Qwen-MICo shows excellent performance on visually complex tasks that demand \textbf{a deep understanding of lighting and optics}, and produces outputs with strong aesthetic appeal.}

  \label{fig:mico-case-3}
\end{figure*}

\begin{figure*}[t]
  \centering
  \includegraphics[width=\linewidth]{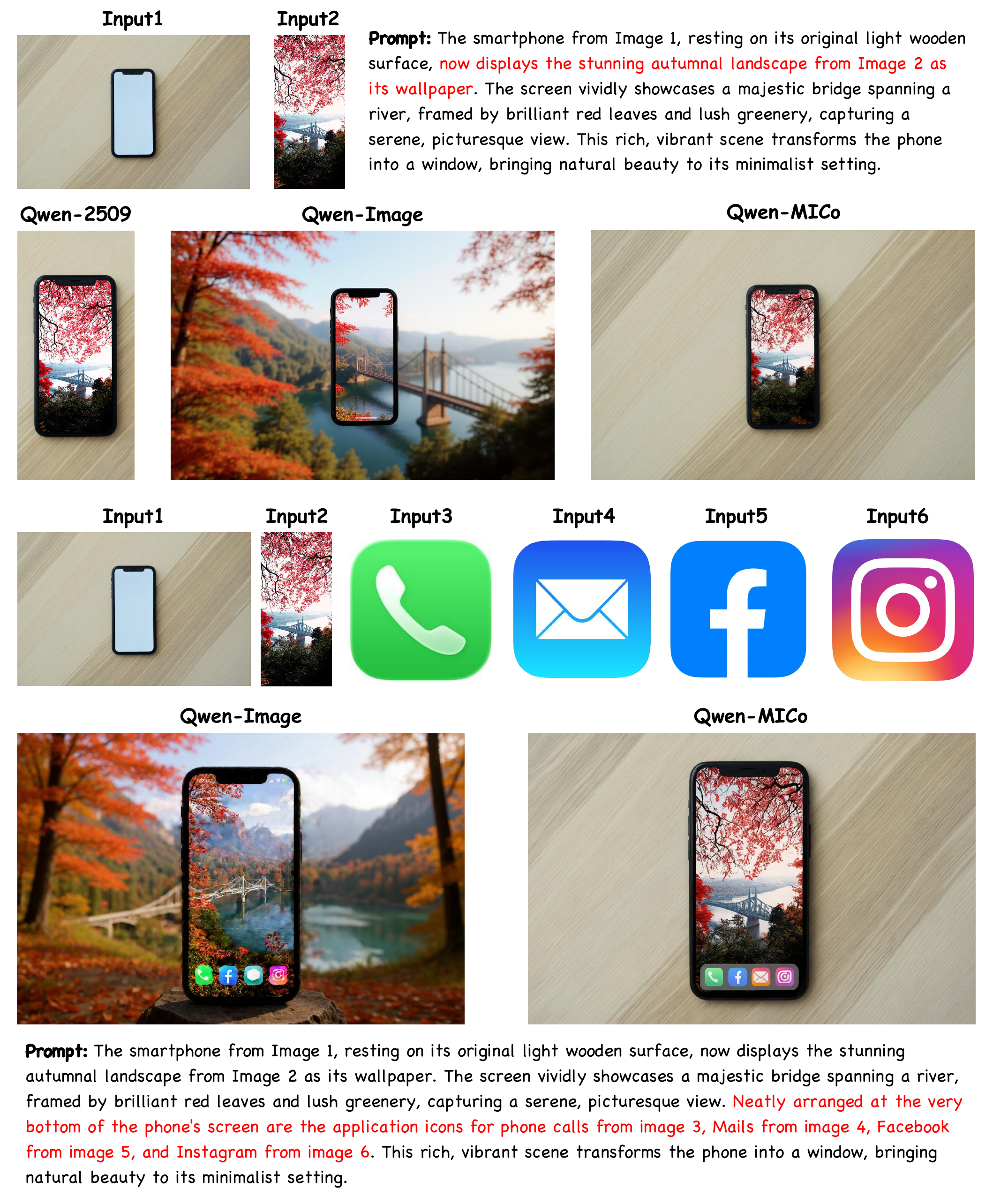}
  \caption{Qwen-MICo preserves the entire appearance of Input~2 while correctly interpreting the prompt phrase \textit{“resting on its original light wooden surface”}. 
On top of this, it supports more image inputs than Qwen-Image-2509, and accurately renders all application icons onto the phone’s home screen.
}

  \label{fig:mico-case-5}
\end{figure*}

In addition, we observe that Qwen-MICo exhibits several remarkable emergent capabilities, with representative examples shown in Fig.~\ref{fig:mico-case-1}–Fig.~\ref{fig:mico-case-5}.


\begin{figure*}[t]
  \centering
  \includegraphics[width=\linewidth]{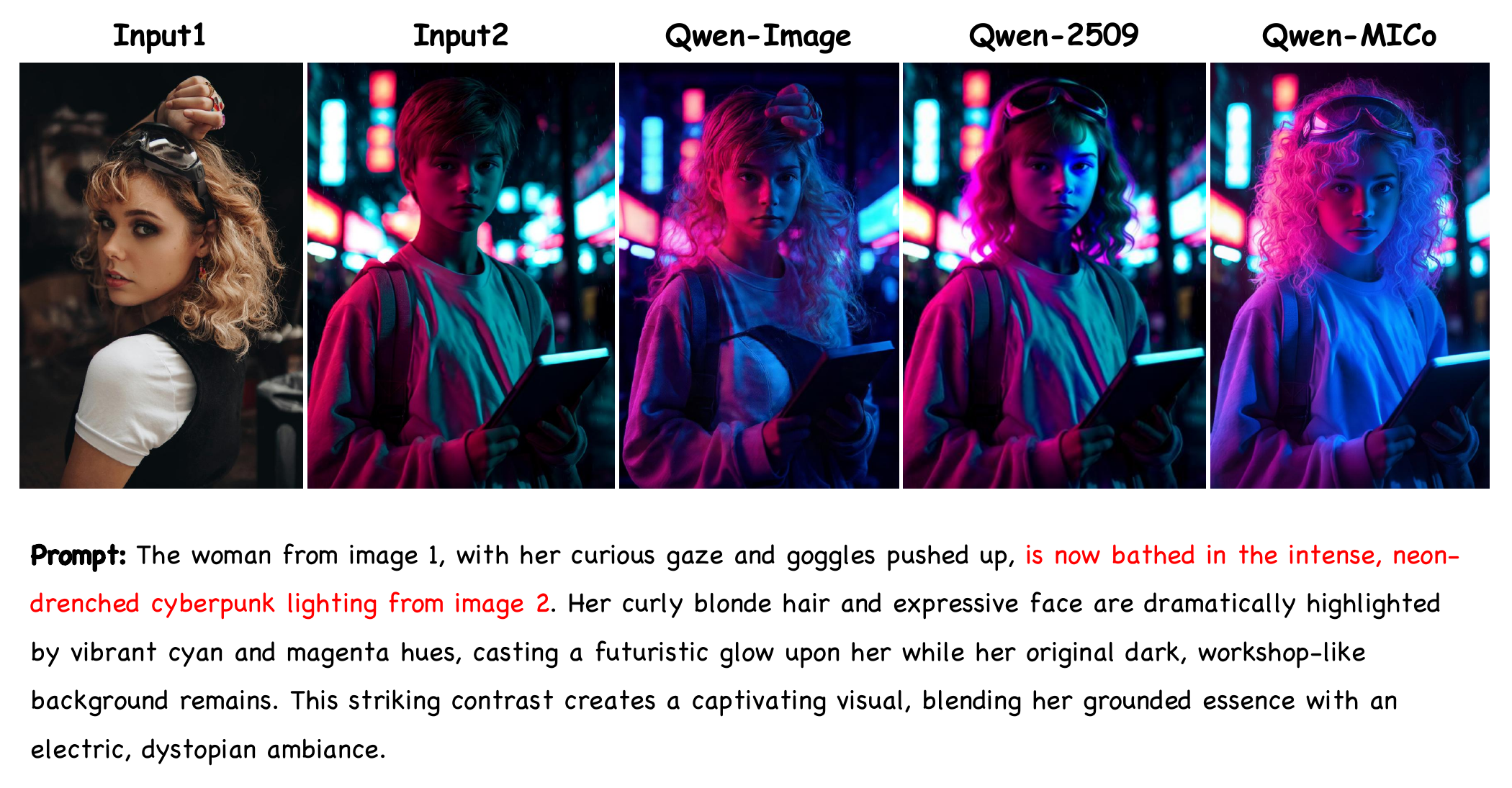}
  \caption{Qwen-MICo preserves the subject’s identity while accurately modeling \textbf{lighting and shading}, transferring the illumination from Image~2 onto the girl in Image~1.
}
\end{figure*}

\clearpage
\twocolumn[%
\vspace{1em}
]

{
    \small
    \bibliographystyle{ieeenat_fullname}
    \bibliography{main}  
}


\end{document}